\documentclass{article} 
\usepackage{iclr2026_conference,times}

\usepackage{amsmath,amsfonts,bm}

\def\eqref#1{equation~\ref{#1}}

\def\1{\bm{1}}

\DeclareMathAlphabet{\mathsfit}{\encodingdefault}{\sfdefault}{m}{sl}
\SetMathAlphabet{\mathsfit}{bold}{\encodingdefault}{\sfdefault}{bx}{n}

\usepackage{hyperref}
\usepackage{url}
\usepackage[utf8]{inputenc} 
\usepackage[T1]{fontenc}    
\usepackage{hyperref}       
\usepackage{url}            
\usepackage{booktabs}       
\usepackage{amsfonts}       
\usepackage{nicefrac}       
\usepackage{microtype}      
\usepackage{xcolor}         
\definecolor{lstbg}{RGB}{248,248,248}
\definecolor{lstframe}{RGB}{220,220,220}

\usepackage{wrapfig}
\usepackage{mathtools}
\usepackage{adjustbox}
\usepackage{tabularx}     
\usepackage{array}        
\newcolumntype{Y}{>{\raggedright\arraybackslash}X} 
\usepackage{enumitem}
\usepackage{amsmath}
\usepackage{epigraph}
\usepackage{soul}
\usepackage{graphicx}
\usepackage{float}
\usepackage{tabularx}
\usepackage{array}
\usepackage{adjustbox}
\usepackage[capitalize,noabbrev]{cleveref}
\usepackage{graphicx}
\usepackage{subcaption}
\usepackage[most]{tcolorbox}  
\usepackage{xcolor} 
\usepackage{listings}
\usepackage{minitoc}
\usepackage{enumitem}
\usepackage{csquotes}
\usepackage{listings}
\usepackage{amssymb}
\usepackage[table]{xcolor}
\usepackage[toc,page,header]{appendix}
\usepackage[dvipsnames]{xcolor}

\lstdefinelanguage{yaml}{
  morekeywords={true,false,null,yes,no},
  sensitive=false,
  morecomment=[l]{\#},
  morestring=[b]",
  morestring=[b]',
}

\lstdefinestyle{prompt}{
  basicstyle=\ttfamily\small,
  breaklines=true,
  frame=single
}

\crefname{lstlisting}{listing}{listings}

\lstdefinestyle{yaml}{
  basicstyle=\ttfamily\footnotesize,
  backgroundcolor=\color{gray!5},
  frame=single,
  breaklines=true,
  columns=fullflexible,
  moredelim=[s][\color{blue}]{:}{ },
  morestring=[b]',
  morestring=[b]",
  comment=[l]{\#},
  keywordstyle=\color{teal}\bfseries,
}

\lstdefinelanguage{json}{
    basicstyle=\ttfamily\footnotesize,
    numbers=none,
    showstringspaces=false,
    breaklines=true,
    frame=single, 
    backgroundcolor=\color{gray!8},
    morestring=[b]",
    literate=
     *{0}{{{\color{blue}0}}}{1}
      {1}{{{\color{blue}1}}}{1}
      {2}{{{\color{blue}2}}}{1}
      {3}{{{\color{blue}3}}}{1}
      {4}{{{\color{blue}4}}}{1}
      {5}{{{\color{blue}5}}}{1}
      {6}{{{\color{blue}6}}}{1}
      {7}{{{\color{blue}7}}}{1}
      {8}{{{\color{blue}8}}}{1}
      {9}{{{\color{blue}9}}}{1}
      {:}{{{\color{red}:}}}{1},
}
\definecolor{crimson}{RGB}{220, 20, 60}

\newcolumntype{P}[1]{>{\centering\arraybackslash}p{#1}}
\hypersetup{
    colorlinks=true,
    linkcolor=orange,
    filecolor=magenta,      
    urlcolor=cyan,
    citecolor=blue,
    pdftitle={eri},
    pdfpagemode=FullScreen,
    }

\definecolor{takeawaycolor}{RGB}{170,220,210}
\colorlet{takeawaycolor}{takeawaycolor!10}
\newcounter{takeawaycounter}
\newenvironment{takeaway}[1][]{%
    \refstepcounter{takeawaycounter}%
    \begin{tcolorbox}[
        enhanced,
        breakable,
        title=#1,
        colback=takeawaycolor,
        colframe=teal!50!black,
        colbacktitle=takeawaycolor,
        fonttitle=\bfseries,
        coltitle=black,
        attach boxed title to top left={yshift=-3mm, xshift=2mm},
        boxed title style={size=small, colback=takeawaycolor, frame hidden},
        sharp corners,
        rounded corners,
        arc=3mm,
        top=2mm,
        bottom=1mm,
        left=1mm,
        right=1mm,
        width=\linewidth+1mm,
    ]%
}{%
    \end{tcolorbox}
}
\iclrfinalcopy 
\title{Hypothesis Hunting with Evolving Networks of Autonomous Scientific Agents}

\author{Tennison Liu$^1$, Silas Ruhrberg Estévez$^1$, David L. Bentley$^2$, Mihaela van der Schaar$^1$ \\
$^1$ DAMTP, University of Cambridge; $^2$ School of Medicine, University of Colorado
}

\begin{document}
\lstdefinestyle{pyclean}{
  language=Python,
  basicstyle=\ttfamily\small,
  numbers=left,
  numberstyle=\scriptsize\color{gray},
  stepnumber=1,
  numbersep=8pt,
  showstringspaces=false,
  breaklines=true,
  frame=single,
  framerule=0.5pt,
  rulecolor=\color{lstframe},
  backgroundcolor=\color{lstbg},
  tabsize=2,
  keepspaces=true
}

\setlist[itemize]{left=0pt,itemsep=2pt,topsep=2pt,parsep=0pt}

\maketitle
\begin{abstract}
Large-scale scientific datasets—spanning health biobanks, cell atlases, Earth reanalyses, and more—create opportunities for exploratory discovery unconstrained by specific research questions. We term this process \textit{hypothesis hunting}: the cumulative search for insight through sustained exploration across vast and complex hypothesis spaces. To support it, we introduce \texttt{AScience}, a framework modeling discovery as the interaction of agents, networks, and evaluation norms, and implement it as \texttt{ASCollab}, a distributed system of LLM-based research agents with heterogeneous behaviors. These agents self-organize into evolving networks, continually producing and peer-reviewing findings under shared standards of evaluation. Experiments show that such social dynamics enable the accumulation of expert-rated results along the diversity–quality–novelty frontier, including rediscoveries of established biomarkers, extensions of known pathways, and proposals of new therapeutic targets. While wet-lab validation remains indispensable, our experiments on cancer cohorts demonstrate that socially structured, agentic networks can sustain exploratory hypothesis hunting at scale.
\end{abstract}

\section{Introduction}

Modern science is increasingly shaped by \textit{large-scale digital snapshots} of the world: biobanks containing millions of genomes and health records \citep{bycroft2018uk}, cell atlases charting tissues at single-cell resolution \citep{regev2017human}, and global reanalysis datasets tracing Earth systems over decades \citep{hersbach2020era5}. These collections, built from sustained large-scale measurement, contain hidden mechanisms, associations, and regularities that remain undiscovered. Systematically probing such datasets for insight defines a new problem setting that we term \textbf{hypothesis hunting}:

\begin{takeaway}[Hypothesis Hunting]
\textit{Hypothesis hunting} is the continuous and diverse exploration of large-scale datasets to surface promising findings that guide subsequent human investigation and experimental validation.
\end{takeaway}

\begin{wrapfigure}[11]{r}{0.45\linewidth} 
    \vspace{-1.1em}
    \centering
    \includegraphics[width=\linewidth]{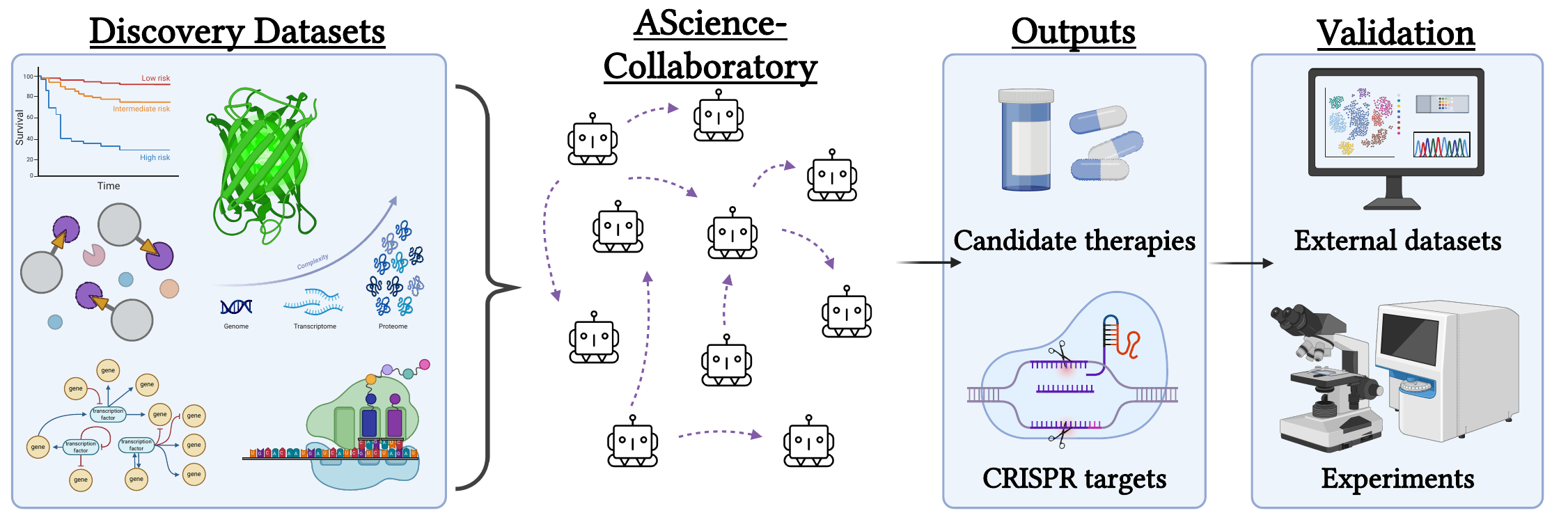}
    \caption{\textbf{Hypothesis hunting.}~Large-scale datasets are explored by autonomous networks of research agents that collaborate, peer-review, and refine findings to surface promising directions for human validation.}
    \label{fig:hypothesis_hunting}
\end{wrapfigure}
This mode of discovery holds vast potential but is limited when pursued by human scientists alone. The obstacles are twofold: \textbf{scale}, with millions of samples and thousands of variables creating a combinatorial explosion of possible analyses; and \textbf{coordination}, since meaningful progress often requires knowledge, tools, and perspectives scattered across disciplines \citep{balietti2015disciplinary}. An autonomous system capable of broad exploration, iterative refinement, and cumulative knowledge building can directly address these challenges, surfacing candidate findings for further human inquiry and wet-lab validation.

Recent advances in autonomous science have begun to make this vision tangible. Of note, large language model (LLM) agents, equipped with tools, domain expertise, and reasoning capabilities, can propose hypotheses, design experiments, execute analyses, and interpret results \citep{lu2024ai,gottweis2025towards}. While representing important advances, these systems are designed around answering \textit{predefined research questions}. Hypothesis hunting, by contrast, imposes more fundamental demands—chief among them requirements for \textbf{exploration}, \textbf{evaluation}, and \textbf{accumulation}. The search space of possible questions and approaches in large-scale datasets is vast yet sparse, calling for diverse and adaptive exploration strategies coupled with mechanisms for knowledge consolidation. Importantly, the potential discoveries vary widely in type and scope (e.g., from biomarker associations to therapeutic leads), making their significance heterogeneous, context dependent, and difficult to assess without the anchor of a specific question. Finally, value derives not only from isolated results but also from accumulation: the incremental refinement, layering, and recombination of discoveries into evolving research programs \citep{lehman2008exploiting}.

Our central insight is that advancing systematic discovery in this setting requires not just autonomous agents but \textbf{networks of agents}, where the social dynamics are crucial to uncovering novel exploratory directions and turning scattered findings into ongoing knowledge accumulation. In human science, progress accelerates when communities of investigators pursue diverse approaches, critique one another’s claims, and cross-pollinate across domains, producing layered bodies of evidence \citep{fortunato2018science}. These cooperative and competitive dynamics are mediated by networks, flows of attention and investigation budgets, and shared frameworks for evaluation. Our central insight is that enhancing this social aspect of agentic systems is key to unlocking hypothesis hunting at scale. 

To formalize this idea, we introduce \texttt{AScience}, a framework that models collective science through four interacting components: (i) an epistemic landscape of possible approaches, (ii) heterogeneous scientific agents, (iii) networks that route attention and collaboration, and (iv) robust evaluation mechanisms of `good science'. We instantiate this framework in \texttt{AScience-Collaboratory} (\texttt{ASCollab}), a distributed system of heterogeneous LLM-based scientific agents that generate and refine diverse findings de novo, interact to form evolving networks, and are guided by shared scientific standards. Through this system, discoveries emerge not from a single agent pursuing a single goal, but from a community engaged in parallel exploration, quality control, and cumulative refinement. 

Empirically, we find that these social dynamics support the continuous accumulation of expert-credible findings along the diversity–quality–novelty frontier. Agents distributed within the network display heterogeneous and evolving behaviors, while collaboration structures reorganize endogenously to drive broader exploration. Applied to three cancer cohorts from \textbf{The Cancer Genome Atlas} \citep{weinstein2013cancer}, integrating transcriptomic, proteomic, pathway, and clinical survival data, \texttt{ASCollab} generates diverse and potentially interesting findings—ranging from \textbf{(1)} rediscoveries of established cancer drivers to \textbf{(2)} extensions of ferroptosis pathways and \textbf{(3)} proposals of new therapeutic targets—showcasing the promise of networked agents for hypothesis hunting at scale.

\textbf{Contributions.}~Our core contributions are three-fold:
\vspace{-0.5em}
\begin{enumerate}[leftmargin=*,itemsep=0.4ex,parsep=-0.1ex,topsep=0.25ex]
    \item \textbf{Framework.}~We formalize hypothesis hunting—the continuous, open-ended exploration of large-scale datasets for promising discoveries—and introduce \texttt{AScience}, a framework capturing the social dynamics of cumulative scientific progress.
    \item \textbf{Agentic system.}~We instantiate this framework as \texttt{ASCollab}, a distributed system of heterogeneous LLM-based research agents that generate, critique, and refine findings through endogenous interaction and shared evaluation standards.
    \item \textbf{Empirical evidence.}~On TCGA, \texttt{ASCollab} sustains cumulative exploration and yields findings judged novel, high-quality, and diverse, spanning validated biomarkers, pathway-level extensions, and new therapeutic hypotheses of potential scientific or clinical significance.
\end{enumerate}
\section{Formalism}
\label{sec:formalism}

\subsection{Modeling the Social Dynamics of Science}

Scientific progress does not unfold as a collection of isolated researchers running analyses, but as a collective process shaped by ideas, agents, interactions, and shared evaluative norms. To capture this, we model science as a dynamic system in which agents navigate an epistemic landscape, exchange information through networks, and adapt to feedback and accumulating knowledge.

\textbf{Datasets.}~We take as provided large-scale datasets $\mathcal{D}$ (e.g.,~genomic cohorts, astronomical surveys), providing the empirical basis from which research questions, methods, and findings are drawn.

\textbf{Epistemic landscape.}~A research field, defined implicitly by $\mathcal{D}$, can be viewed as an \emph{epistemic landscape}: a structured space of possible \textit{approaches}, each with some intrinsic scientific value \citep{weisberg2009epistemic}. Conceptually, approaches differ in the questions they pose, the instruments and analytic methods they use, and the theoretical framings they adopt. Formally, let $\mathcal{X}$ denote the space of approaches, with $x \in \mathcal{X}$ indexing a specific approach, and let $\mathcal{Y}\subseteq \mathbb{R}$ denote epistemic significance. The landscape is defined by a ground-truth mapping $f:\mathcal{X}\to\mathcal{Y}$, and is generally rugged: some approaches yield high significance (local peaks), others little insight (valleys), with global maxima representing approaches closest to the set of underlying truths encoded in $\mathcal{D}$.

\textbf{Perceived epistemic significance.}~Agents do not observe $f$ directly. Instead, they form beliefs about a time-varying \emph{perceived landscape} $\tilde f_t$. This perception is shaped by the history of visible outputs $H_t \subseteq \mathcal{O}$, the network of attention $W_t$, and shared standards of evaluation $I$. Conceptually, $\tilde f_t = \Gamma_t\big(f;,I,,W_t,,H_t\big)$, aggregating the influence of prior findings, diffusion through networks, and evaluation standards. Importantly, $\tilde f_t$ evolves even if $f$ is fixed: a finding of high intrinsic value loses perceived significance once it becomes common knowledge and judged non-novel via $I$.

\textbf{Scientific agents.}~Researchers or research groups are modeled as heterogeneous agents $a^i \in \mathcal{A}={1,2,\ldots,N}$, each with a state vector $a_t^i = (x_t^i,\theta_t^i,e_t^i,b_t^i,\rho_t^i)$:
\vspace{-0.75em}
\begin{enumerate}[leftmargin=*,itemsep=0.4ex,parsep=-0.1ex]
    \item $x_t^i$: current \textit{approach} (coordinates on the landscape);
    \item $\theta_t^i$: \textit{epistemic behavior} (e.g., explore vs.~exploit, collaborate vs.~solo, risk-taking vs.~conservative);
    \item $e_t^i$: \textit{expertise} (or specialization within the research field);
    \item $b_t^i$: \textit{belief state} (summarizing the agent’s internal view of the field);
    \item $\rho_t^i$: publicly visible history such as publications or citations (collectively termed \textit{reputation}).
\end{enumerate}
\vspace{-0.75em}
Then, each agent can be viewed as following a stochastic research policy $x_{t+1} \sim \pi(\cdot \mid x_t, \theta_t, e_t, b_t)$ to produce research outputs $o_t^i \in \mathcal{O}$.

\textbf{Networks of agents.}~Social interactions (e.g., information sharing, collaboration) are modeled as a time-varying weighted directed graph $G_t = (\mathcal{A}, W_t)$, where $W_t = (w_{ij}^t){i,j\in\mathcal{A}}$ and each edge $w{ij}^t$ captures the \textit{attention} agent $a_t^i$ allocates to signals from agent $a_t^j$ (in particular, $\rho_t^j$). These interactions shape belief states $b_t^i$, which in turn guide agents’ subsequent strategies of research.

\textbf{Standards of evaluation.}~Collective progress also depends on shared standards $I$ that define what counts as valuable science. Formally, $I$ comprises: (i) an evaluation operator $\Xi_t$ mapping each output to a score $s_t^i = \Xi_t(o_t^i;\tilde f_t)$ (e.g., novelty, rigor, reproducibility), and (ii) a consequence operator $\rho_{t+1}^i \leftarrow \Upsilon_t(o_t^i, s_t^i, \rho_t^i)$ mapping outputs and scores to updates of $\rho_t^i$ (e.g., reputational gains through publication or citation). These standards govern visibility and guide how resources and attention flow.

Together, the perceived landscape $\tilde f_t$, agent states ${a_t^i}$, networks $G_t$, outputs $H_t$, and standards $I$ co-evolve. Agents adapt strategies to new information; networks reorganize as attention shifts; evaluation influences perceived significance. Social research dynamics thus emerge from feedback among agents, ideas, networks, and norms.

\subsection{Problem Setting}

The formalism above is general, but different scientific contexts emphasize different dynamics of landscapes, networks, and evaluation. We distinguish two broad settings:
\vspace{-0.5em}
\begin{enumerate}[leftmargin=*,itemsep=0.4ex,parsep=-0.1ex]
    \item \textbf{Goal-driven.}~In this setting, agents converge on approaches aimed at a narrow objective (e.g., identifying an antibody for a novel pathogen). Progress is measured by how quickly and reliably the target is reached. Once the optimal solution is known, further rediscoveries add little beyond verification or robustness. These scenarios have clear endpoints and natural stopping rules.
    \item \textbf{Cumulative.}~Here, agents explore a broad topic (e.g., cancer biology) through diverse questions, methods, and perspectives. Individual research episodes are partly independent yet mutually enabling: results accumulate, tools are repurposed, and findings open new lines of inquiry. Progress has no natural endpoint but unfolds as layered evidence that reshapes the field.
\end{enumerate}
\vspace{-0.5em}
The focus of \textit{hypothesis hunting} is squarely on the second setting, characterized by open-ended exploration without objectives; heterogeneous findings whose value is context- and time-dependent; and the dynamic evolution of perceived significance, collaboration networks, and research directions.
\section{Evolving Networks of Autonomous Scientific Agents}

\begin{figure}[t!]
    \vspace{-2em}
    \centering
    \includegraphics[width=0.99\linewidth]{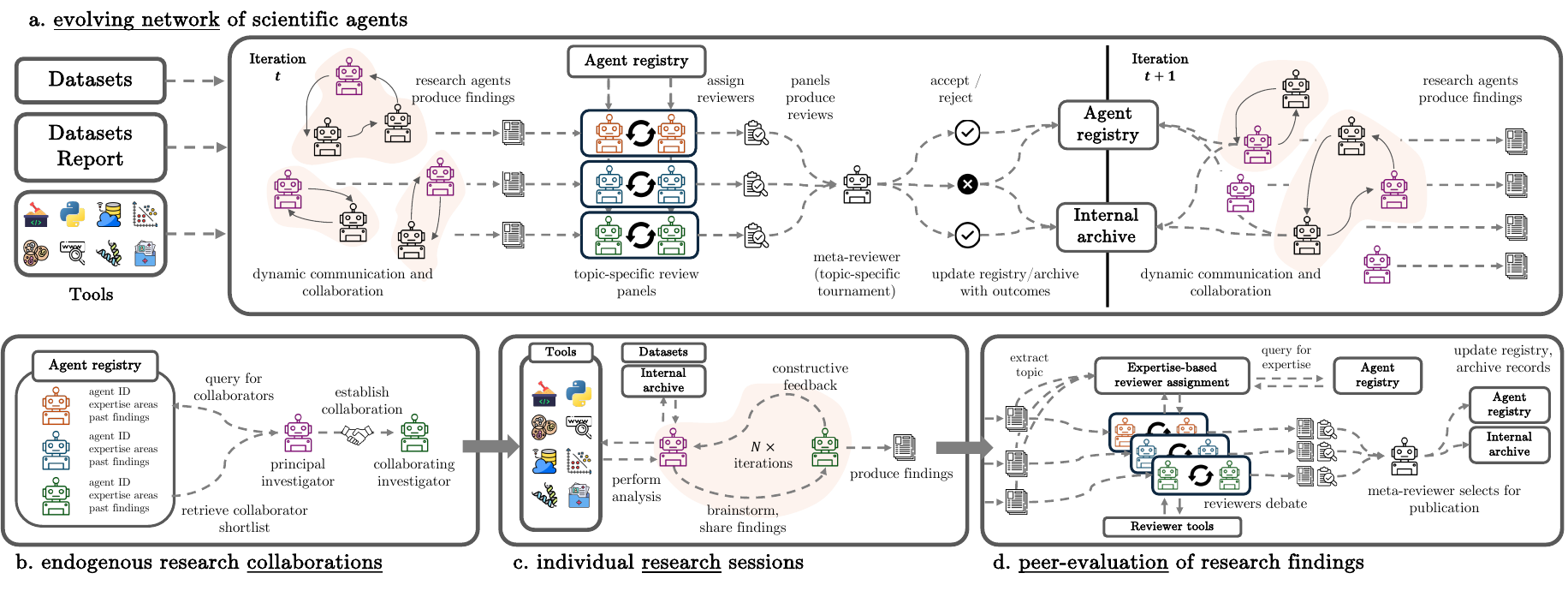}
    \caption{
    \textbf{\texttt{ASCollab}.}~Evolving network of distributed agents hypothesis hunting.
    }
    \label{fig:system_overview}
    \vspace{-1em}
\end{figure}

In this section, we instantiate the \texttt{AScience} framework as \texttt{AScience-Collaboratory} (\texttt{ASCollab}), a system designed to support hypothesis hunting over large-scale datasets $\mathcal{D}$. \texttt{ASCollab} consists of a heterogeneous population of scientific agents—differing in expertise, epistemic behavior, and reputational status—embedded in an evolving network. Agents independently pursue research, but also collaborate, and peer-review each other’s findings. Crucially, the network itself is not fixed: patterns of collaboration and attention emerge endogenously from agent capabilities and evolving histories. An overview of the system is shown in \Cref{fig:system_overview}.

\subsection{Agent Network Infrastructure}

To enable such network dynamics, \texttt{ASCollab} maintains two shared-memory structures that serve as the system’s connective tissue: (i) an \textbf{agent registry}, which tracks the active research community, and (ii) an \textbf{internal archive}, which stores the body of accumulated research outputs. Together, these structures allow agents to locate relevant collaborators, access prior findings, and update their internal beliefs. Conceptually, they play a role similar to academic infrastructure such as Google Scholar or PubMed: supporting both the discovery of collaborators and the retrieval of relevant literature.

\textbf{Agent registry.}~The registry maintains the public profiles of all agents in $\mathcal{A}$, indexed by unique identifiers. Each entry contains: (i) a profile of the agent’s expertise $e^i$; and (ii) reputation metadata $\rho^i$, including the number of accepted papers and citations received. Specifically, agents are prompted periodically to update $e_t^i$ based on their recent research. Reputation metadata is updated by the system, reflecting findings accepted into the internal archive and accumulated citations.

\textbf{Internal archive.}~The archive functions as the entire network’s publication record, containing all outputs accepted into the network. Each record is indexed by a unique paper identifier and stores rich metadata: authoring agents (including collaborators), title, abstract, full manuscript, associated code, public reviews, bibliography, and time of acceptance. The data archive is automatically updated at the end of each research round, as papers accepted through network peer-review are added to the archive. The exact schema of the two structures are described in \Cref{app:schema}.

\textbf{Query mechanics.}~Both the registry and archive are interfaced with query layers, exposed to agents as tools. Queries are resolved via vector search \citep{salton1975vector}, with embeddings tailored to the underlying content: agent expertise representations (derived from $e^i$) for the registry, and title–abstract embeddings for the archive. This enables natural text queries such as ``expertise in pathway analysis'' or papers investigating ``certain pathways in renal carcinoma.'' Retrieval-augmented generation (RAG) integrates these results directly into agent reasoning, allowing artefacts of the shared memory to shape research trajectories, collaboration choices, and review judgments \citep{lewis2020retrieval}.

\subsection{Heterogeneous Scientific Agents}

To encourage sustained exploration in \texttt{ASCollab}, we introduce heterogeneity across agents rather than assigning them uniform roles. Without such diversity, agents risk converging too quickly on similar solutions, limiting the coverage of the research landscape \citep{hong2004groups}. By varying epistemic behavior and expertise, the system maintains a broader range of strategies and perspectives, which supports more balanced exploration and exploitation.

In principle, heterogeneity can be introduced through many mechanisms, including specialist training, distinct underlying LLMs, or access to different toolkits. In \texttt{ASCollab}, we focus on two dimensions: \emph{epistemic behavior} ($\theta^i$) and \emph{expertise} ($e^i$). At system initialization, we query the underlying LLM to generate a set of distinct behavioral profiles and areas of expertise, which are embedded in each agent’s system prompt, akin to assigning a scientific \textit{persona} \citep{park2023generative}.

\textbf{Epistemic behavior.}~Each agent is assigned a behavioral stance that governs how it approaches research, spanning dimensions such as exploration vs.~exploitation and independence vs.~collaboration (see Appendix~\ref{app:epistemic_profiles} for the full set of stances). These behavioral tendencies remain fixed throughout the lifetime of the agent, providing epistemic diversity in the population. \textbf{Expertise.}~Expertise profiles are sampled with respect to the dataset $\mathcal{D}$. For example, when working with TCGA cohorts, sampled expertise includes capabilities such as differential expression analysis, gene set enrichment, or drug–target interaction analysis. Unlike epistemic behavior, expertise is adaptive and periodically updated by agents to reflect latest specialization \citep{lazer2007network}.

\textbf{Memory.}~In contrast to public artefacts in the shared archive, each agent maintains a private memory of its past work, including findings not accepted into the archive \citep{wu2024autogen}. Agents can query this memory to retrieve prior findings or intermediate code analyses, enabling continuity and reuse in their research programs. Together, epistemic behavior, expertise, and memory define each agent's research policy $\pi^i$, shaping how it selects problems, collaborates, and produces findings over time.

\subsection{Collaboration and Research Sessions}

Research in \texttt{ASCollab} unfolds through distributed sessions (or rounds), in which each agent acts as a primary investigator tasked with producing a new finding. Importantly, each agent is free to determine its own research plan, with no pre-specified workflows or constraints.

\textbf{Research environment.}~Agents have direct access to the datasets $\mathcal{D}$ and operate within identical, but dedicated computational environment. This environment provides a suite of tools: (i) query interfaces to the agent registry, internal archive, private memory, and external literature search; (ii) collaboration mechanisms for identifying, inviting, and exchanging messages with other agents; and (iii) a sandbox for executing code, preloaded with domain-specific software relevant to the dataset (e.g., differential expression analysis, pathway enrichment, or survival modeling in the case of TCGA cohorts).

\textbf{Reasoning loop.}~Agents plan their research activity via the \texttt{ReAct} framework \citep{yao2023react}, cycling through three steps: plan and reason $\rightarrow$ act by invoking tools or writing code $\rightarrow$ observe resulting outcome (see \Cref{app:reasoning_tool_use} for details). Each research session consists of up to $M$ such iterations, though agents determine dynamically how to allocate reasoning across exploration, analysis, or collaboration. \textbf{Collaboration model.}~Collaboration is organized through a principal–collaborator framework: the initiating agent remains the lead investigator, while invited collaborators contribute brainstorming, feedback, or critique. Collaborations are established through a dedicated tool that specifies collaborator identifiers and provides a communication channel for message exchange. At the conclusion of a session, each agent produces a standardized research report (see \Cref{app:agent_prompts}) summarizing findings, evidence, and references (from both external sources and the internal archive). Any code written during the session is automatically extracted. Thus, each output $o_t$ takes the form of a (report, code) pair, and with $N$ agents, each round yields $N$ such outputs.

\subsection{Evaluation via Peer-Review}

The final component is the protocol of evaluation $I$, which we design through a structured peer-review process. This provides a collective input for assessing the quality of outputs and controlling which findings enter the archive. Specifically, the evaluation mechanism consists of two stages:

\textbf{Review stage.}~Each research output $o_t^i = (\text{report}, \text{code})$ is assigned to a panel of $K$ reviewers. Reviewers are selected by querying the agent registry with the title and metadata of the submission to identify agents with relevant expertise, ensuring that the authoring agent is excluded. The process is double-blind, and an agent may serve on multiple review panels concurrently. Reviewers provide structured assessments (see Appendix~\ref{app:agent_prompts}), scoring the submission on a $1$–$4$ scale along four dimensions: (i) \emph{support} (empirical and logical grounding of claims), (ii) \emph{soundness} (technical rigor), (iii) \emph{significance} (contribution to advancing knowledge), and (iv) \emph{originality} (novelty of ideas, methods, or results). Specifically, reviewers cannot execute code, but they have visibility of the complete codebase as well as query tools for the archive and literature to contextualize evaluation.

\textbf{Meta-review stage.}~Following the review stage, submissions are clustered thematically, and each cluster is assigned to a meta-reviewing agent. Unlike research/review agents, the meta-reviewer is a dedicated agent whose role is to execute a tournament consisting of related submissions \citep{goldberg1991comparative}. Given $L$ submissions and their associated reviews, the meta-reviewer produces a relative judgment of merit: assigning each paper a score on a $0$–$1$ scale, together with a brief written justification. To calibrate decisions, the meta-reviewer is also shown randomly sampled reference papers from the archive. By design, the meta-reviewer does not access external tools, relying solely on its reasoning and the provided reviews. \textbf{Acceptance.}~The combined review and meta-review scores form the evaluation operator $\Xi_t$, yielding a vector of scores for all outputs. The top $1/K$ fraction of outputs produced by the network in each round is accepted into the internal archive, becoming part of the network’s shared memory. Citations within accepted papers are propagated to update archival entries and agent metadata in the registry, reflecting reputational gains. This consequence operator $\Upsilon_t$ closes the evaluation loop by mapping outputs and scores into visible signals on individual findings and agents and by propagating statistics through the archive and registry.

Each round of research therefore concludes with evaluation and acceptance updates, after which agents continue their research with an updated registry and archive. Over $T$ rounds, this feedback loop ensures that the network's collective behavior is continually shaped by cumulative findings.
\vspace{-0.5em}
\section{Related Works}

Our work is primarily related to three lines of research (for an extended survey, please see \Cref{app:extended_rws}).

\textbf{Data-driven discovery.}~Classical approaches focus on deriving hypotheses directly from empirical data. These include \textit{symbolic regression}, which recovers closed-form equations \citep{schmidt2009distilling, brunton2016discovering, udrescu2020ai}; logic programming and rule discovery, which extract relational or propositional hypotheses \citep{quinlan1990learning, clark1989cn2, lin2020generalized}; and \textit{causal discovery}, which infers causal graphs from observational data using independence tests, scoring criteria, or functional assumptions \citep{spirtes2000causation, zheng2018dags, peters2014causal}.

\textbf{LLM-augmented discovery.}~Recent work has explored replacing handcrafted inductive biases with the scientific priors encoded in large language models. LLMs are deployed as \textit{search operators}, generating and modifying candidate hypotheses—often expressed in code—guided by evaluators such as solvers, experiments, or reward signals. This paradigm has enabled advances in algorithm and mathematical discovery \citep{romera2024mathematical, novikov2025alphaevolve}, and has been applied across domains including neural architecture search \citep{chen2023evoprompting}, decision trees \citep{liu2025decision}, symbolic equations \citep{shojaee2025llmsr}, theorem proving \citep{trinh2024solving}, robotics reward design \citep{ma2024eureka}, and molecular design \citep{wang2025efficient}, underscoring the potential for LLM-based search to broaden and accelerate discovery.

\textbf{Agentic science.}~An emerging direction concerns agentic systems that integrates LLMs with tool-rich, memory-augmented agents to automate aspects of the scientific process. One line emphasizes automating experimental workflows, e.g., chemical synthesis or biomedical pipelines \citep{m2024augmenting, ruan2024automatic, huang2025biomni, qu2025crispr}. More directly relevant are systems for hypothesis generation and refinement, such as the \texttt{AI Scientist} \citep{lu2024ai}, which can autonomously generate ideas, run analyses, and draft papers, and the \texttt{AI Co-Scientist} \citep{gottweis2025towards}, which employs multi-agent debate and evolution to refine hypotheses. Related work on automated falsification \citep{huang2025automated} and domain-specific instantiations \citep{saeedi2025astroagents, ghafarollahi2025sciagents} further illustrate this paradigm.
\section{Experiments}

We evaluate \texttt{ASCollab} on three hypothesis hunting tasks in cancer genomics.

\textbf{Large-scale datasets.}~We use The Cancer Genome Atlas (TCGA) \citep{weinstein2013cancer}, a landmark initiative that molecularly characterized over $20,000$ tumor and matched normal samples across $33$ cancer types, producing multi-omic datasets that have underpinned thousands of studies \citep{tomczak2015review}. TCGA is a prime testbed for hypothesis hunting for three reasons: (i) \emph{real-world impact}, as uncovering new mechanisms, biomarkers, and therapeutic targets in cancer remains a major scientific and clinical challenge; (ii) \emph{scale and richness}, as TCGA provides comprehensive molecular measurements across many cancers, with numerous yet-unexplored associations and potential insights; and (iii) \emph{reproducibility}, as TCGA is an open-access resource.

We focus on three cohorts: kidney renal clear cell carcinoma (\textbf{KIRC}) \citep{cancer2013comprehensive}, diffuse large B-cell lymphoma (\textbf{DLBC}) \citep{weinstein2013cancer}, and pancreatic adenocarcinoma (\textbf{PAAD}) \citep{raphael2017integrated}. For each cohort, we integrate (1) bulk RNA-sequencing, (2) protein expression arrays, (3) clinical phenotypes, (4) survival outcomes, together with (5) pathway annotations and (6) drug–target information from the Probes \& Drugs database \citep{skuta2017probes}. We do not apply any preprocessing to these datasets. Full dataset details are provided in \Cref{app:dataset_details}. Beyond providing the datasets, we do not specify concrete research question, instead instructing the agents to \textit{`discover novel, strongly supported, and scientifically significant findings on the provided datasets'}.

\textbf{Evaluation.}~Evaluating autonomous scientific systems is inherently challenging, as outputs are less predictable, open-ended, and heterogeneous \citep{lu2024ai,gottweis2025towards}. We assess findings along three complementary dimensions: (1) \emph{Novelty}: the extent to which a finding introduces ideas or associations not already present in the literature; (2) \emph{Quality}: the rigor, plausibility, and evidential support of the finding; (3) \emph{Diversity}: the breadth of the hypothesis space covered.

\textbf{Implementation details.}~We deploy a population of $N=16$ agents for $T=40$ rounds. Each research session is capped at $M=40$ \texttt{ReAct} loops, with $K=2$ reviewers assigned per paper and meta-review tournaments of size $L=4$. All agents use \texttt{gpt-4o-2024-08-06} \citep{hurst2024gpt} as the underlying LLM (knowledge cut-off: 	October 2023), with \texttt{text-embedding-3-small} for retrieval-augmented queries. Multi-agent orchestration is implemented via \texttt{LangGraph}, and agent sandboxes run in isolated environments on a \texttt{32-core AMD Epyc Milan 7713 CPU}. Additional implementation details are provided in \Cref{app:additional_details}.

\begin{figure}[t]
    \vspace{-2em}
    \centering
    \includegraphics[width=0.99\linewidth]{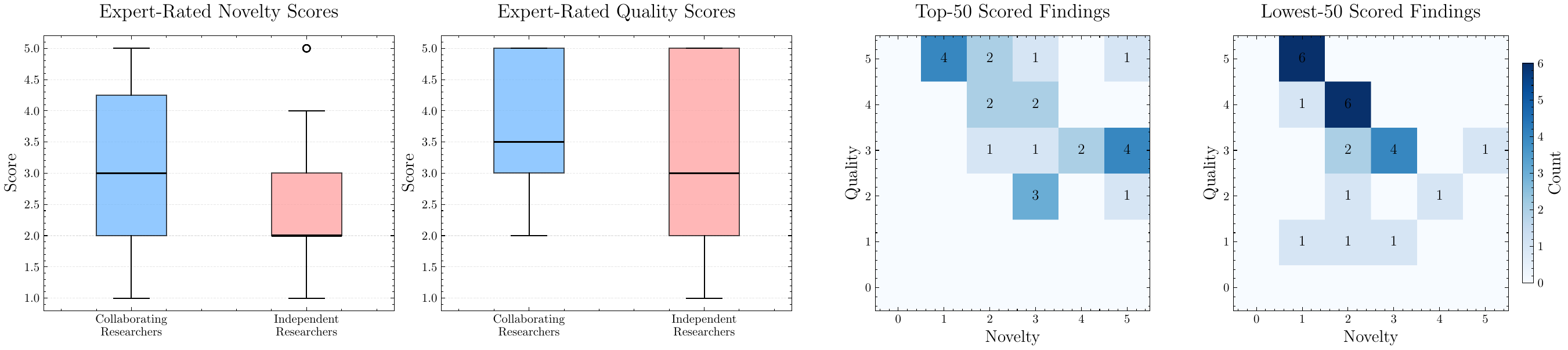}
    \includegraphics[width=0.99\linewidth]{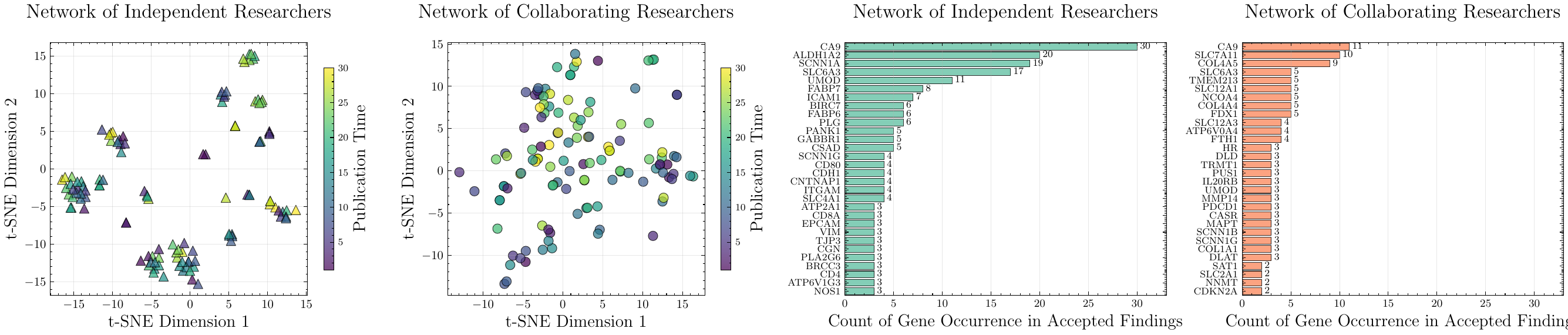}
    \caption{\textbf{Evaluation of novelty, quality, and diversity of findings produced by research network}.}
    \label{fig:novelty_quality_diversity}
    \vspace{-2em}
\end{figure}

\subsection{Evaluation of Produced Findings}

To assess the effectiveness of \texttt{ASCollab}, we compare it against an ablated baseline where agents operate independently. These agents retain the same hyperparameters and computational budget as the network but lack access to global data stores (agent registry and internal archive). Consequently, each can rely only on its own past outputs, with no possibility of collaboration or cross-pollination.

\textbf{Evaluation protocol.}
From both settings, we select the top $25$ outputs as ranked by meta-review scores. These outputs are then evaluated by a domain expert (using a rubric described in \Cref{app:evaluation_rubric}): \emph{Novelty} ($1$–$5$): from $1$ = essentially already published in the same form (including analyses), to $5$ = substantial novel contribution with no prior precedent. \emph{Quality} ($1$–$5$): from $1$ = conflicting with strong established evidence, to $5$ = highly plausible, well-supported by related literature, or generalizable across datasets or cancer types. For \emph{diversity}, we analyze the distribution of implicated gene targets and compute embedding-based visualization of abstracts.

\textbf{Results.}~Results of the expert evaluation are shown in \Cref{fig:novelty_quality_diversity}. Expert evaluation indicates that findings produced by \texttt{ASCollab} are both more novel and of higher quality than those from independent agents. In the baseline, many findings were near-duplicates, with almost half overlapping substantially. Consequently, a filtering step was required to ensure $25$ unique findings. In contrast, \texttt{ASCollab} outputs were more heterogeneous, with no duplication in the top $25$ findings.

Embedding visualizations of research findings via t-SNE \citep{maaten2008visualizing} reveal that independent agents tend to converge (over time) on a narrow set of areas, whereas \texttt{ASCollab} agents explore outward into a broader space of hypotheses. Gene-level histograms corroborate this pattern: independent agents concentrate heavily on a small subset of targets, while \texttt{ASCollab} produces findings implicating a wider range of genes. Finally, novelty–quality frontiers show that the highest-scoring outputs from \texttt{ASCollab} also received the strongest expert ratings. Taken together, \texttt{ASCollab}, by leveraging social dynamics and shared memory, sustains cumulative exploration that yields discoveries which are not only more diverse, but also consistently of higher quality and novelty.

\subsection{Detailed Case Studies}
Beyond aggregate evaluation, two domain experts examined a subset of findings in depth. Here we highlight three representative findings, with full reports, analyses, and reproducible code in \Cref{app:agentic_case_studies}. For balance, we also include negative cases where the peer-review pipeline recommended rejection, illustrating how the system filters overlap with prior literature or unsupported claims.

\begin{takeaway}[Multi-gene Ferroptosis axis in KIRC (\Cref{case_study_1})]
Agents identified a ferroptosis module involving \texttt{ACSL4}, \texttt{GPX4}, and \texttt{FTH1} in kidney cancer, a part of which was later independently discovered and published in \citet{Zheng2025} (after knowledge cut-off of LLM, and manual examination of research trace revealed this work was not retrieved by agent). This finding, supported by DepMap essentiality data and prior mixed evidence \citep{Guo2015,Huang2019fth1,Zou2019gpx4}, was enabled by the primary agent extending earlier findings by another agent (on \texttt{SLC7A11}/\texttt{ALOX5}) into a broader mechanistic hypothesis.
\end{takeaway}

\begin{takeaway}[SLC5A2 and ABCC8 in PAAD (\Cref{case_study_2})]
Agents proposed \texttt{SLC5A2} (SGLT2) and \texttt{ABCC8} as therapeutic targets in pancreatic adenocarcinoma, anticipating a July 2025 publication that independently confirmed the \texttt{SLC5A2–PAAD} link \citep{Xie2025}. This finding, contextualized against prior work emphasizing \texttt{SGLT1} \citep{Du2022} and largely non-oncologic studies of \texttt{SGLT2} \citep{Jurczak2011}, illustrates how agent collaboration surfaced a novel target class while situating results within the transporter literature.
\end{takeaway}

\begin{takeaway}[BIRC5 validation and PRKD1 extension in KIRC (\Cref{case_study_3})]
Agents independently reproduced the established role of \texttt{BIRC5} (Survivin) as a diagnostic and prognostic marker in KIRC \citep{Wang2021birc}, strengthening confidence by re-deriving results from scratch on TCGA data. Building on this, collaboration extended the analysis to implicate \texttt{PRKD1} as a putative tumor-suppressive regulator, proposing complementary therapeutic leads.
\end{takeaway}

\begin{figure}[t]
    \vspace{-2em}
    \centering
    \includegraphics[width=0.99\linewidth]{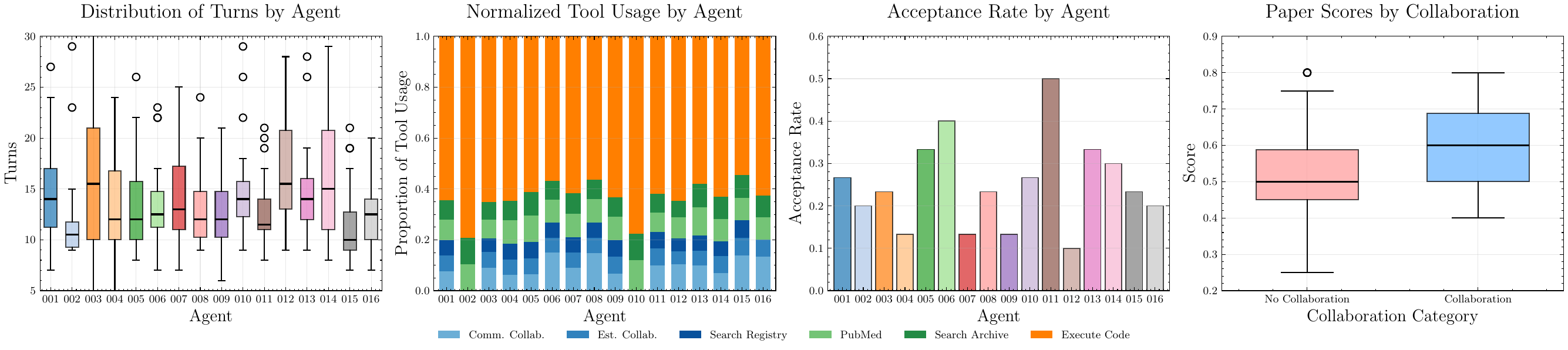}
    \begin{subfigure}{0.49\linewidth}
        \centering
        \includegraphics[width=\linewidth]{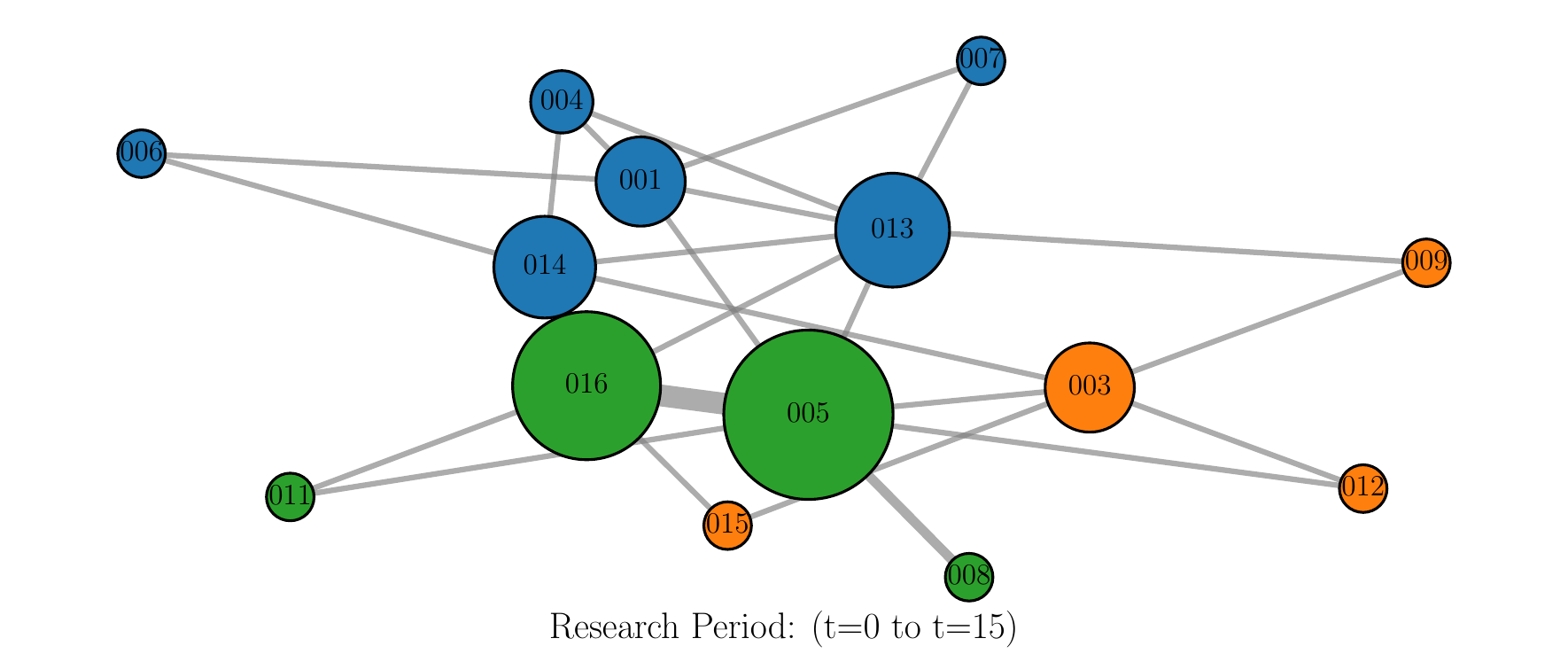}
    \end{subfigure}
    \hfill
    \begin{subfigure}{0.49\linewidth}
        \centering
        \includegraphics[width=\linewidth]{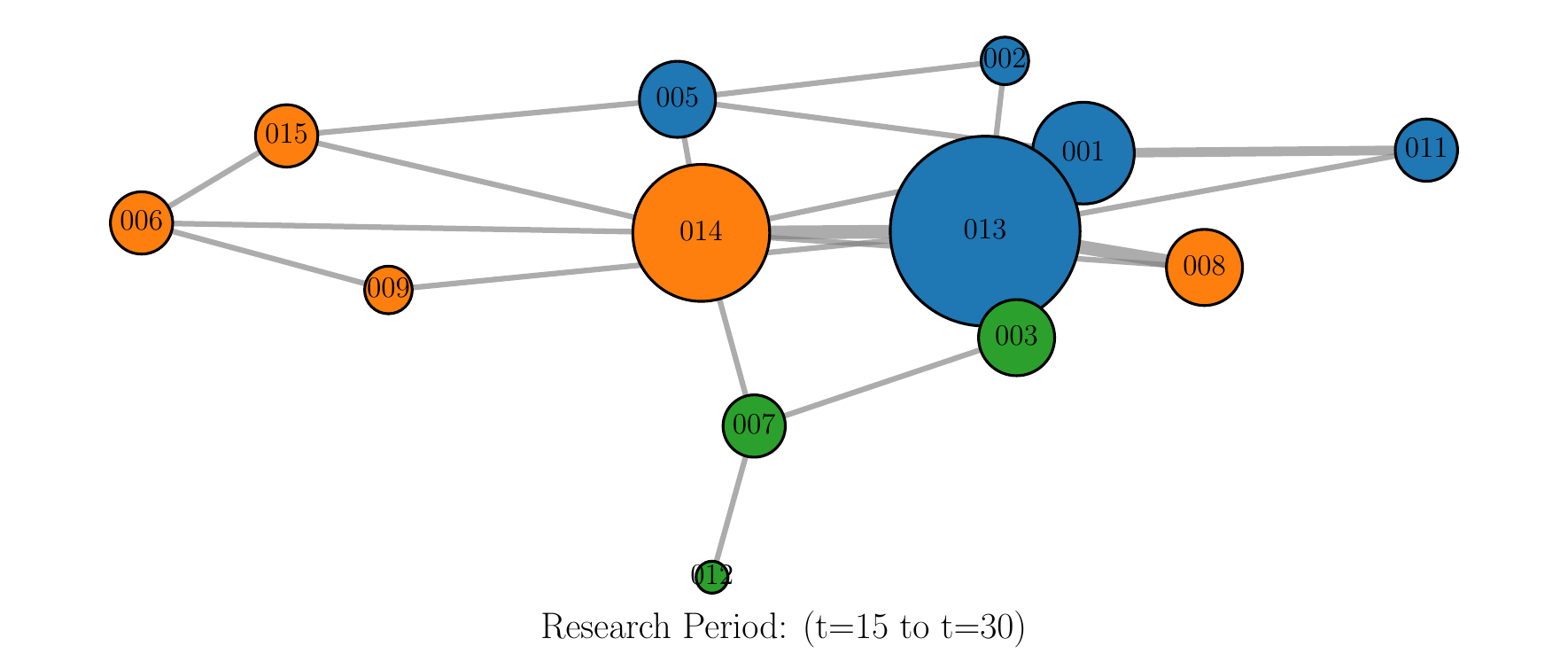}
    \end{subfigure}
    \caption{\textbf{Heterogeneous agent behaviors and endogenous network evolution.}}
    \label{fig:heterogeneous_behaviour}
    \vspace{-1.5em}
\end{figure}

\subsection{Agent Behaviors and Network Evolution}

To investigate how heterogeneity and social dynamics emerge in \texttt{ASCollab}, we examine (i) diversity in epistemic behavior across agents and (ii) the temporal evolution of collaboration networks.

\textbf{Heterogeneous epistemic behaviors.}~In \Cref{fig:heterogeneous_behaviour}, we visualize distributions of session lengths and normalized tool usage aggregated across research sessions. Agents display marked differences in research style: some (e.g., \texttt{agent\_002}, \texttt{agent\_015}) conduct very lean research, while others pursue considerably longer investigations. Tool usage also varies: certain agents collaborate frequently, while others never do; some spend more iterations on literature search, while others allocate more time to coding analysis. Notably, outputs produced through collaboration receive systematically higher meta-review scores than those produced in isolation, despite the double-blind evaluation process, underscoring the epistemic value of collaborations.

\textbf{Dynamic collaboration networks.}~Collaboration patterns also evolve endogenously over time. Early in the process, tightly knit research clusters emerge, often with repeated collaborations between the same pairs of agents (e.g., \texttt{agent\_016} and \texttt{agent\_005}). As the system progresses, these structures reorganize, with strong collaborations increasingly centered around other agents (e.g., \texttt{agent\_013}), indicating reorganization as the network adapts to emerging areas of inquiry.

\begin{figure}[h] 
    \centering \includegraphics[width=0.99\linewidth]{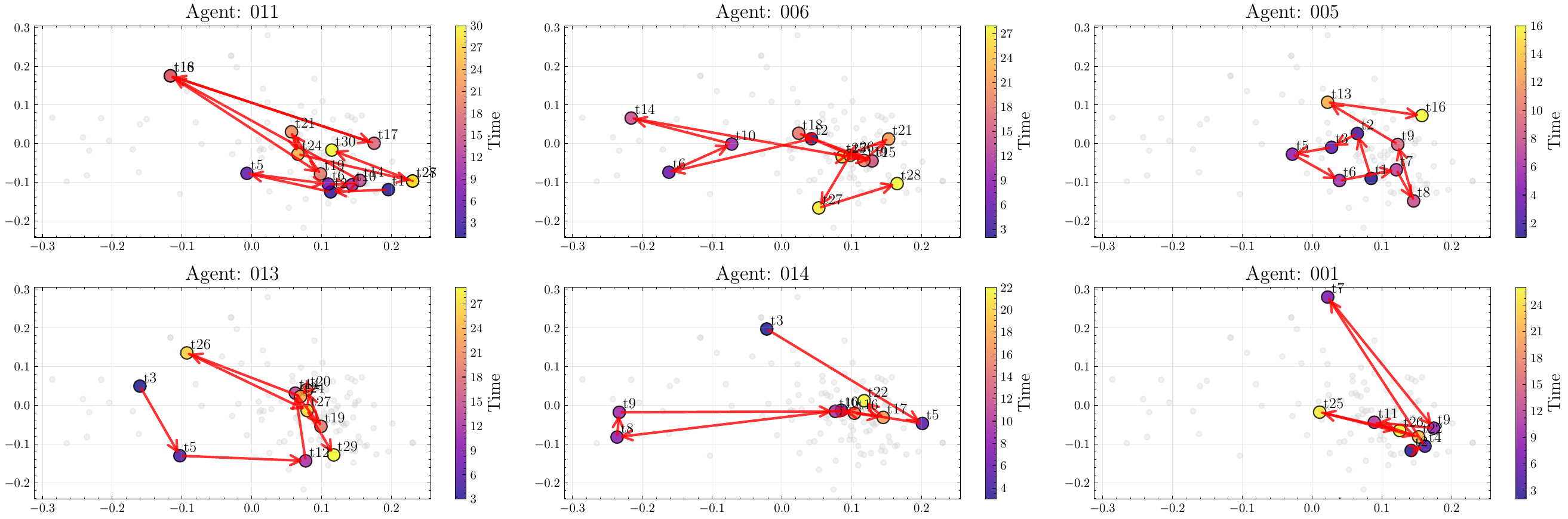} 
    \caption{\textbf{Exploration trajectory of heterogeneous agents.}} 
    \label{fig:exploration_behavior} 
    
\end{figure}

\textbf{Distinct exploratory trajectories.}~To further probe individual behavior, we visualize the research trajectories of the six most productive agents in \Cref{fig:exploration_behavior}. Clear research tendencies emerge: some agents prefer local refinement and exploitation, repeatedly developing variations of an idea, while others adopt a more exploratory stance, testing hypotheses across multiple modalities and directions, underscoring diverse strategies that enable breadth and depth in hypothesis hunting.
        
\section{Discussion}

In closing, this work investigates \textit{hypothesis hunting} as a new problem setting for autonomous discovery and instantiated it in \texttt{ASCollab}, a network of heterogeneous scientific agents whose social dynamics enable cumulative exploration. Across three cancer cohorts in TCGA, we found that \texttt{ASCollab} produces findings that are diverse, and rated as higher in novelty and quality than comparable system of independent agents, underscoring the importance of endogenous communication between distributed agents, evolving under social dynamics. \textbf{Future works.}~At the same time, our claims should be interpreted with care: results are demonstrated within genomics, and generalization to other domains remains to be established; expert-based evaluation of novelty and quality, while structured, is inevitably subjective; and current experiments operate with modest agent populations and a single LLM backbone. Most importantly, \textcolor{Maroon}{\textbf{findings represent candidate hypotheses rather than validated biomedical discoveries}}, and experimental validation is required before translational impact can be claimed. These direction highlight the promise of networked autonomous agents as a catalyst to accelerate and broaden the frontier of scientific inquiry, surfacing diverse, high-quality hypotheses as a preface to human investigations.

\clearpage
\bibliography{iclr2026_conference}

\begin{thebibliography}{70}
\providecommand{\natexlab}[1]{#1}
\providecommand{\url}[1]{\texttt{#1}}
\expandafter\ifx\csname urlstyle\endcsname\relax
  \providecommand{\doi}[1]{doi: #1}\else
  \providecommand{\doi}{doi: \begingroup \urlstyle{rm}\Url}\fi

\bibitem[Balietti et~al.(2015)Balietti, M{\"a}s, and Helbing]{balietti2015disciplinary}
Stefano Balietti, Michael M{\"a}s, and Dirk Helbing.
\newblock On disciplinary fragmentation and scientific progress.
\newblock \emph{PloS one}, 10\penalty0 (3):\penalty0 e0118747, 2015.

\bibitem[Battelli \& Cho(2011)Battelli and Cho]{Battelli2011}
Chiara Battelli and Daniel~C Cho.
\newblock mtor inhibitors in renal cell carcinoma.
\newblock \emph{Therapy}, 8\penalty0 (4):\penalty0 359–367, July 2011.
\newblock ISSN 1744-831X.
\newblock \doi{10.2217/thy.11.32}.
\newblock URL \url{http://dx.doi.org/10.2217/thy.11.32}.

\bibitem[Brunton et~al.(2016)Brunton, Proctor, and Kutz]{brunton2016discovering}
Steven~L Brunton, Joshua~L Proctor, and J~Nathan Kutz.
\newblock Discovering governing equations from data by sparse identification of nonlinear dynamical systems.
\newblock \emph{Proceedings of the national academy of sciences}, 113\penalty0 (15):\penalty0 3932--3937, 2016.

\bibitem[Bycroft et~al.(2018)Bycroft, Freeman, Petkova, Band, Elliott, Sharp, Motyer, Vukcevic, Delaneau, O’Connell, et~al.]{bycroft2018uk}
Clare Bycroft, Colin Freeman, Desislava Petkova, Gavin Band, Lloyd~T Elliott, Kevin Sharp, Allan Motyer, Damjan Vukcevic, Olivier Delaneau, Jared O’Connell, et~al.
\newblock The uk biobank resource with deep phenotyping and genomic data.
\newblock \emph{Nature}, 562\penalty0 (7726):\penalty0 203--209, 2018.

\bibitem[Cervenkova et~al.(2023)Cervenkova, Palek, Moulisova, Liska, Daum, Mohelnikova-Duchonova, and Soucek]{Cervenkova2023}
Lenka Cervenkova, Richard Palek, Vladimira Moulisova, Vaclav Liska, Ondrej Daum, Beatrice Mohelnikova-Duchonova, and Pavel Soucek.
\newblock Protein expression and localization of abc transporters in pancreatic adenocarcinoma: Prognostic role of abcc8.
\newblock \emph{Pancreatology}, 23\penalty0 (8):\penalty0 978–987, December 2023.
\newblock ISSN 1424-3903.
\newblock \doi{10.1016/j.pan.2023.10.008}.
\newblock URL \url{http://dx.doi.org/10.1016/j.pan.2023.10.008}.

\bibitem[Chen et~al.(2023)Chen, Dohan, and So]{chen2023evoprompting}
Angelica Chen, David Dohan, and David So.
\newblock Evoprompting: Language models for code-level neural architecture search.
\newblock In \emph{Thirty-seventh Conference on Neural Information Processing Systems}, 2023.
\newblock URL \url{https://openreview.net/forum?id=ifbF4WdT8f}.

\bibitem[Clark \& Niblett(1989)Clark and Niblett]{clark1989cn2}
Peter Clark and Tim Niblett.
\newblock The cn2 induction algorithm.
\newblock \emph{Machine learning}, 3\penalty0 (4):\penalty0 261--283, 1989.

\bibitem[Dorigo \& Gambardella(1997)Dorigo and Gambardella]{dorigo1997ant}
Marco Dorigo and Luca~Maria Gambardella.
\newblock Ant colonies for the travelling salesman problem.
\newblock \emph{biosystems}, 43\penalty0 (2):\penalty0 73--81, 1997.

\bibitem[Du et~al.(2022)Du, Gu, Deng, Kong, Guo, Jin, Bao, Fu, and Li]{Du2022}
Jiali Du, Jichun Gu, Junyuan Deng, Lei Kong, Yujie Guo, Chen Jin, Yun Bao, Deliang Fu, and Ji~Li.
\newblock The expression and survival significance of sodium glucose transporters in pancreatic cancer.
\newblock \emph{BMC Cancer}, 22\penalty0 (1), January 2022.
\newblock ISSN 1471-2407.
\newblock \doi{10.1186/s12885-021-09060-4}.
\newblock URL \url{http://dx.doi.org/10.1186/s12885-021-09060-4}.

\bibitem[El-Mokadem et~al.(2014)El-Mokadem, Fitzpatrick, Rai, Cunningham, Pratt, Fleming, and Nabi]{ElMokadem2014}
Ismail El-Mokadem, John Fitzpatrick, Bhavan Rai, J.~Cunningham, Norman Pratt, Stewart Fleming, and Ghulam Nabi.
\newblock Significance of chromosome 9p status in renal cell carcinoma: A systematic review and quality of the reported studies.
\newblock \emph{BioMed Research International}, 2014:\penalty0 1–11, 2014.
\newblock ISSN 2314-6141.
\newblock \doi{10.1155/2014/521380}.
\newblock URL \url{http://dx.doi.org/10.1155/2014/521380}.

\bibitem[Fortunato et~al.(2018)Fortunato, Bergstrom, B{\"o}rner, Evans, Helbing, Milojevi{\'c}, Petersen, Radicchi, Sinatra, Uzzi, et~al.]{fortunato2018science}
Santo Fortunato, Carl~T Bergstrom, Katy B{\"o}rner, James~A Evans, Dirk Helbing, Sta{\v{s}}a Milojevi{\'c}, Alexander~M Petersen, Filippo Radicchi, Roberta Sinatra, Brian Uzzi, et~al.
\newblock Science of science.
\newblock \emph{Science}, 359\penalty0 (6379):\penalty0 eaao0185, 2018.

\bibitem[Gao et~al.(2024)Gao, Lan, Li, Yuan, Ding, Zhou, Xu, and Li]{gao2024large}
Chen Gao, Xiaochong Lan, Nian Li, Yuan Yuan, Jingtao Ding, Zhilun Zhou, Fengli Xu, and Yong Li.
\newblock Large language models empowered agent-based modeling and simulation: A survey and perspectives.
\newblock \emph{Humanities and Social Sciences Communications}, 11\penalty0 (1):\penalty0 1--24, 2024.

\bibitem[Ghafarollahi \& Buehler(2025)Ghafarollahi and Buehler]{ghafarollahi2025sciagents}
Alireza Ghafarollahi and Markus~J Buehler.
\newblock Sciagents: automating scientific discovery through bioinspired multi-agent intelligent graph reasoning.
\newblock \emph{Advanced Materials}, 37\penalty0 (22):\penalty0 2413523, 2025.

\bibitem[Goldberg \& Deb(1991)Goldberg and Deb]{goldberg1991comparative}
David~E Goldberg and Kalyanmoy Deb.
\newblock A comparative analysis of selection schemes used in genetic algorithms.
\newblock In \emph{Foundations of genetic algorithms}, volume~1, pp.\  69--93. Elsevier, 1991.

\bibitem[Gottweis et~al.(2025)Gottweis, Weng, Daryin, Tu, Palepu, Sirkovic, Myaskovsky, Weissenberger, Rong, Tanno, et~al.]{gottweis2025towards}
Juraj Gottweis, Wei-Hung Weng, Alexander Daryin, Tao Tu, Anil Palepu, Petar Sirkovic, Artiom Myaskovsky, Felix Weissenberger, Keran Rong, Ryutaro Tanno, et~al.
\newblock Towards an ai co-scientist.
\newblock \emph{arXiv preprint arXiv:2502.18864}, 2025.

\bibitem[Guo et~al.(2015)Guo, German, Bai, Barnes, Guo, Qi, Lou, Liang, Jonasch, Mills, and Ding]{Guo2015}
Huifang Guo, Peter German, Shanshan Bai, Sean Barnes, Wei Guo, Xiangjie Qi, Hongxiang Lou, Jiyong Liang, Eric Jonasch, Gordon~B. Mills, and Zhiyong Ding.
\newblock The pi3k/akt pathway and renal cell carcinoma.
\newblock \emph{Journal of Genetics and Genomics}, 42\penalty0 (7):\penalty0 343–353, July 2015.
\newblock ISSN 1673-8527.
\newblock \doi{10.1016/j.jgg.2015.03.003}.
\newblock URL \url{http://dx.doi.org/10.1016/j.jgg.2015.03.003}.

\bibitem[Hersbach et~al.(2020)Hersbach, Bell, Berrisford, Hirahara, Hor{\'a}nyi, Mu{\~n}oz-Sabater, Nicolas, Peubey, Radu, Schepers, et~al.]{hersbach2020era5}
Hans Hersbach, Bill Bell, Paul Berrisford, Shoji Hirahara, Andr{\'a}s Hor{\'a}nyi, Joaqu{\'\i}n Mu{\~n}oz-Sabater, Julien Nicolas, Carole Peubey, Raluca Radu, Dinand Schepers, et~al.
\newblock The era5 global reanalysis.
\newblock \emph{Quarterly journal of the royal meteorological society}, 146\penalty0 (730):\penalty0 1999--2049, 2020.

\bibitem[Hong \& Page(2004)Hong and Page]{hong2004groups}
Lu~Hong and Scott~E Page.
\newblock Groups of diverse problem solvers can outperform groups of high-ability problem solvers.
\newblock \emph{Proceedings of the National Academy of Sciences}, 101\penalty0 (46):\penalty0 16385--16389, 2004.

\bibitem[Huang et~al.(2019)Huang, Qiu, Huang, Zhou, Zhou, and Luo]{Huang2019fth1}
Huimei Huang, Yuyun Qiu, Guilian Huang, Xiaohui Zhou, Xiaoying Zhou, and Wenqi Luo.
\newblock Value of ferritin heavy chain (fth1) expression in diagnosis and prognosis of renal cell carcinoma.
\newblock \emph{Medical Science Monitor}, 25:\penalty0 3700–3715, May 2019.
\newblock ISSN 1643-3750.
\newblock \doi{10.12659/msm.914162}.
\newblock URL \url{http://dx.doi.org/10.12659/MSM.914162}.

\bibitem[Huang et~al.(2025{\natexlab{a}})Huang, Jin, Li, Li, Cand{\`e}s, and Leskovec]{huang2025automated}
Kexin Huang, Ying Jin, Ryan Li, Michael~Y Li, Emmanuel Cand{\`e}s, and Jure Leskovec.
\newblock Automated hypothesis validation with agentic sequential falsifications.
\newblock \emph{arXiv preprint arXiv:2502.09858}, 2025{\natexlab{a}}.

\bibitem[Huang et~al.(2025{\natexlab{b}})Huang, Zhang, Wang, Qu, Lu, Roohani, Li, Qiu, Li, Zhang, et~al.]{huang2025biomni}
Kexin Huang, Serena Zhang, Hanchen Wang, Yuanhao Qu, Yingzhou Lu, Yusuf Roohani, Ryan Li, Lin Qiu, Gavin Li, Junze Zhang, et~al.
\newblock Biomni: A general-purpose biomedical ai agent.
\newblock \emph{bioRxiv}, pp.\  2025--05, 2025{\natexlab{b}}.

\bibitem[Hurst et~al.(2024)Hurst, Lerer, Goucher, Perelman, Ramesh, Clark, Ostrow, Welihinda, Hayes, Radford, et~al.]{hurst2024gpt}
Aaron Hurst, Adam Lerer, Adam~P Goucher, Adam Perelman, Aditya Ramesh, Aidan Clark, AJ~Ostrow, Akila Welihinda, Alan Hayes, Alec Radford, et~al.
\newblock Gpt-4o system card.
\newblock \emph{arXiv preprint arXiv:2410.21276}, 2024.

\bibitem[Jurczak et~al.(2011)Jurczak, Lee, Birkenfeld, Jornayvaz, Frederick, Pongratz, Zhao, Moeckel, Samuel, Whaley, Shulman, and Kibbey]{Jurczak2011}
Michael~J. Jurczak, Hui-Young Lee, Andreas~L. Birkenfeld, Francois~R. Jornayvaz, David~W. Frederick, Rebecca~L. Pongratz, Xiaoxian Zhao, Gilbert~W. Moeckel, Varman~T. Samuel, Jean~M. Whaley, Gerald~I. Shulman, and Richard~G. Kibbey.
\newblock Sglt2 deletion improves glucose homeostasis and preserves pancreatic b-cell function.
\newblock \emph{Diabetes}, 60\penalty0 (3):\penalty0 890–898, February 2011.
\newblock ISSN 1939-327X.
\newblock \doi{10.2337/db10-1328}.
\newblock URL \url{http://dx.doi.org/10.2337/db10-1328}.

\bibitem[Kennedy \& Eberhart(1995)Kennedy and Eberhart]{kennedy1995particle}
James Kennedy and Russell Eberhart.
\newblock Particle swarm optimization.
\newblock In \emph{Proceedings of ICNN'95-international conference on neural networks}, volume~4, pp.\  1942--1948. ieee, 1995.

\bibitem[Lazer \& Friedman(2007)Lazer and Friedman]{lazer2007network}
David Lazer and Allan Friedman.
\newblock The network structure of exploration and exploitation.
\newblock \emph{Administrative science quarterly}, 52\penalty0 (4):\penalty0 667--694, 2007.

\bibitem[Lehman et~al.(2008)Lehman, Stanley, et~al.]{lehman2008exploiting}
Joel Lehman, Kenneth~O Stanley, et~al.
\newblock Exploiting open-endedness to solve problems through the search for novelty.
\newblock In \emph{ALIFE}, pp.\  329--336, 2008.

\bibitem[Lewis et~al.(2020)Lewis, Perez, Piktus, Petroni, Karpukhin, Goyal, K{\"u}ttler, Lewis, Yih, Rockt{\"a}schel, et~al.]{lewis2020retrieval}
Patrick Lewis, Ethan Perez, Aleksandra Piktus, Fabio Petroni, Vladimir Karpukhin, Naman Goyal, Heinrich K{\"u}ttler, Mike Lewis, Wen-tau Yih, Tim Rockt{\"a}schel, et~al.
\newblock Retrieval-augmented generation for knowledge-intensive nlp tasks.
\newblock \emph{Advances in neural information processing systems}, 33:\penalty0 9459--9474, 2020.

\bibitem[Lin et~al.(2020)Lin, Zhong, Hu, Rudin, and Seltzer]{lin2020generalized}
Jimmy Lin, Chudi Zhong, Diane Hu, Cynthia Rudin, and Margo Seltzer.
\newblock Generalized and scalable optimal sparse decision trees.
\newblock In \emph{International conference on machine learning}, pp.\  6150--6160. PMLR, 2020.

\bibitem[Liu et~al.(2025)Liu, Huynh, and van~der Schaar]{liu2025decision}
Tennison Liu, Nicolas Huynh, and Mihaela van~der Schaar.
\newblock Decision tree induction through {LLM}s via semantically-aware evolution.
\newblock In \emph{The Thirteenth International Conference on Learning Representations}, 2025.
\newblock URL \url{https://openreview.net/forum?id=UyhRtB4hjN}.

\bibitem[Lu et~al.(2024)Lu, Lu, Lange, Foerster, Clune, and Ha]{lu2024ai}
Chris Lu, Cong Lu, Robert~Tjarko Lange, Jakob Foerster, Jeff Clune, and David Ha.
\newblock The ai scientist: Towards fully automated open-ended scientific discovery.
\newblock \emph{arXiv preprint arXiv:2408.06292}, 2024.

\bibitem[M.~Bran et~al.(2024)M.~Bran, Cox, Schilter, Baldassari, White, and Schwaller]{m2024augmenting}
Andres M.~Bran, Sam Cox, Oliver Schilter, Carlo Baldassari, Andrew~D White, and Philippe Schwaller.
\newblock Augmenting large language models with chemistry tools.
\newblock \emph{Nature Machine Intelligence}, 6\penalty0 (5):\penalty0 525--535, 2024.

\bibitem[Ma et~al.(2024)Ma, Liang, Wang, Huang, Bastani, Jayaraman, Zhu, Fan, and Anandkumar]{ma2024eureka}
Yecheng~Jason Ma, William Liang, Guanzhi Wang, De-An Huang, Osbert Bastani, Dinesh Jayaraman, Yuke Zhu, Linxi Fan, and Anima Anandkumar.
\newblock Eureka: Human-level reward design via coding large language models.
\newblock In \emph{The Twelfth International Conference on Learning Representations}, 2024.
\newblock URL \url{https://openreview.net/forum?id=IEduRUO55F}.

\bibitem[Maaten \& Hinton(2008)Maaten and Hinton]{maaten2008visualizing}
Laurens van~der Maaten and Geoffrey Hinton.
\newblock Visualizing data using t-sne.
\newblock \emph{Journal of machine learning research}, 9\penalty0 (Nov):\penalty0 2579--2605, 2008.

\bibitem[Minsky(1986)]{minsky1986society}
Marvin Minsky.
\newblock \emph{Society of mind}.
\newblock Simon and Schuster, 1986.

\bibitem[Motzer et~al.(2013)Motzer, Hutson, Cella, Reeves, Hawkins, Guo, Nathan, Staehler, de~Souza, Merchan, Boleti, Fife, Jin, Jones, Uemura, De~Giorgi, Harmenberg, Wang, Sternberg, Deen, McCann, Hackshaw, Crescenzo, Pandite, and Choueiri]{Motzer2013}
Robert~J. Motzer, Thomas~E. Hutson, David Cella, James Reeves, Robert Hawkins, Jun Guo, Paul Nathan, Michael Staehler, Paul de~Souza, Jaime~R. Merchan, Ekaterini Boleti, Kate Fife, Jie Jin, Robert Jones, Hirotsugu Uemura, Ugo De~Giorgi, Ulrika Harmenberg, Jinwan Wang, Cora~N. Sternberg, Keith Deen, Lauren McCann, Michelle~D. Hackshaw, Rocco Crescenzo, Lini~N. Pandite, and Toni~K. Choueiri.
\newblock Pazopanib versus sunitinib in metastatic renal-cell carcinoma.
\newblock \emph{New England Journal of Medicine}, 369\penalty0 (8):\penalty0 722–731, August 2013.
\newblock ISSN 1533-4406.
\newblock \doi{10.1056/nejmoa1303989}.
\newblock URL \url{http://dx.doi.org/10.1056/NEJMoa1303989}.

\bibitem[Motzer et~al.(2015)Motzer, Escudier, McDermott, George, Hammers, Srinivas, Tykodi, Sosman, Procopio, Plimack, Castellano, Choueiri, Gurney, Donskov, Bono, Wagstaff, Gauler, Ueda, Tomita, Schutz, Kollmannsberger, Larkin, Ravaud, Simon, Xu, Waxman, and Sharma]{Motzer2015}
Robert~J. Motzer, Bernard Escudier, David~F. McDermott, Saby George, Hans~J. Hammers, Sandhya Srinivas, Scott~S. Tykodi, Jeffrey~A. Sosman, Giuseppe Procopio, Elizabeth~R. Plimack, Daniel Castellano, Toni~K. Choueiri, Howard Gurney, Frede Donskov, Petri Bono, John Wagstaff, Thomas~C. Gauler, Takeshi Ueda, Yoshihiko Tomita, Fabio~A. Schutz, Christian Kollmannsberger, James Larkin, Alain Ravaud, Jason~S. Simon, Li-An Xu, Ian~M. Waxman, and Padmanee Sharma.
\newblock Nivolumab versus everolimus in advanced renal-cell carcinoma.
\newblock \emph{New England Journal of Medicine}, 373\penalty0 (19):\penalty0 1803–1813, November 2015.
\newblock ISSN 1533-4406.
\newblock \doi{10.1056/nejmoa1510665}.
\newblock URL \url{http://dx.doi.org/10.1056/NEJMoa1510665}.

\bibitem[Network(2013)]{cancer2013comprehensive}
TCGA~Research Network.
\newblock Comprehensive molecular characterization of clear cell renal cell carcinoma.
\newblock \emph{Nature}, 499\penalty0 (7456):\penalty0 43--49, 2013.

\bibitem[Novikov et~al.(2025)Novikov, V{\~u}, Eisenberger, Dupont, Huang, Wagner, Shirobokov, Kozlovskii, Ruiz, Mehrabian, et~al.]{novikov2025alphaevolve}
Alexander Novikov, Ng{\^a}n V{\~u}, Marvin Eisenberger, Emilien Dupont, Po-Sen Huang, Adam~Zsolt Wagner, Sergey Shirobokov, Borislav Kozlovskii, Francisco~JR Ruiz, Abbas Mehrabian, et~al.
\newblock Alphaevolve: A coding agent for scientific and algorithmic discovery.
\newblock \emph{arXiv preprint arXiv:2506.13131}, 2025.

\bibitem[Park et~al.(2023)Park, O'Brien, Cai, Morris, Liang, and Bernstein]{park2023generative}
Joon~Sung Park, Joseph O'Brien, Carrie~Jun Cai, Meredith~Ringel Morris, Percy Liang, and Michael~S Bernstein.
\newblock Generative agents: Interactive simulacra of human behavior.
\newblock In \emph{Proceedings of the 36th annual acm symposium on user interface software and technology}, pp.\  1--22, 2023.

\bibitem[Peters et~al.(2014)Peters, Mooij, Janzing, and Sch{\"o}lkopf]{peters2014causal}
Jonas Peters, Joris~M Mooij, Dominik Janzing, and Bernhard Sch{\"o}lkopf.
\newblock Causal discovery with continuous additive noise models.
\newblock \emph{The Journal of Machine Learning Research}, 15\penalty0 (1):\penalty0 2009--2053, 2014.

\bibitem[Qu et~al.(2025)Qu, Huang, Yin, Zhan, Liu, Yin, Cousins, Johnson, Wang, Shah, et~al.]{qu2025crispr}
Yuanhao Qu, Kaixuan Huang, Ming Yin, Kanghong Zhan, Dyllan Liu, Di~Yin, Henry~C Cousins, William~A Johnson, Xiaotong Wang, Mihir Shah, et~al.
\newblock Crispr-gpt for agentic automation of gene-editing experiments.
\newblock \emph{Nature Biomedical Engineering}, pp.\  1--14, 2025.

\bibitem[Quinlan(1990)]{quinlan1990learning}
J.~Ross Quinlan.
\newblock Learning logical definitions from relations.
\newblock \emph{Machine learning}, 5\penalty0 (3):\penalty0 239--266, 1990.

\bibitem[Raphael et~al.(2017)Raphael, Hruban, Aguirre, Moffitt, Yeh, Stewart, Robertson, Cherniack, Gupta, Getz, et~al.]{raphael2017integrated}
Benjamin~J Raphael, Ralph~H Hruban, Andrew~J Aguirre, Richard~A Moffitt, Jen~Jen Yeh, Chip Stewart, A~Gordon Robertson, Andrew~D Cherniack, Manaswi Gupta, Gad Getz, et~al.
\newblock Integrated genomic characterization of pancreatic ductal adenocarcinoma.
\newblock \emph{Cancer cell}, 32\penalty0 (2):\penalty0 185--203, 2017.

\bibitem[Regev et~al.(2017)Regev, Teichmann, Lander, Amit, Benoist, Birney, Bodenmiller, Campbell, Carninci, Clatworthy, et~al.]{regev2017human}
Aviv Regev, Sarah~A Teichmann, Eric~S Lander, Ido Amit, Christophe Benoist, Ewan Birney, Bernd Bodenmiller, Peter Campbell, Piero Carninci, Menna Clatworthy, et~al.
\newblock The human cell atlas.
\newblock \emph{elife}, 6:\penalty0 e27041, 2017.

\bibitem[Romera-Paredes et~al.(2024)Romera-Paredes, Barekatain, Novikov, Balog, Kumar, Dupont, Ruiz, Ellenberg, Wang, Fawzi, et~al.]{romera2024mathematical}
Bernardino Romera-Paredes, Mohammadamin Barekatain, Alexander Novikov, Matej Balog, M~Pawan Kumar, Emilien Dupont, Francisco~JR Ruiz, Jordan~S Ellenberg, Pengming Wang, Omar Fawzi, et~al.
\newblock Mathematical discoveries from program search with large language models.
\newblock \emph{Nature}, 625\penalty0 (7995):\penalty0 468--475, 2024.

\bibitem[Ruan et~al.(2024)Ruan, Lu, Xu, He, Chen, Zhang, Xuan, Pan, Fang, Gao, et~al.]{ruan2024automatic}
Yixiang Ruan, Chenyin Lu, Ning Xu, Yuchen He, Yixin Chen, Jian Zhang, Jun Xuan, Jianzhang Pan, Qun Fang, Hanyu Gao, et~al.
\newblock An automatic end-to-end chemical synthesis development platform powered by large language models.
\newblock \emph{Nature communications}, 15\penalty0 (1):\penalty0 10160, 2024.

\bibitem[Saeedi et~al.(2025)Saeedi, Buckner, Aponte, and Aghazadeh]{saeedi2025astroagents}
Daniel Saeedi, Denise~K. Buckner, Jose~C. Aponte, and Amirali Aghazadeh.
\newblock Astroagents: A multi-agent {AI} for hypothesis generation from mass spectrometry data.
\newblock In \emph{Towards Agentic AI for Science: Hypothesis Generation, Comprehension, Quantification, and Validation}, 2025.
\newblock URL \url{https://openreview.net/forum?id=1WUCSNAjjB}.

\bibitem[Salton et~al.(1975)Salton, Wong, and Yang]{salton1975vector}
Gerard Salton, Anita Wong, and Chung-Shu Yang.
\newblock A vector space model for automatic indexing.
\newblock \emph{Communications of the ACM}, 18\penalty0 (11):\penalty0 613--620, 1975.

\bibitem[Schmidt \& Lipson(2009)Schmidt and Lipson]{schmidt2009distilling}
Michael Schmidt and Hod Lipson.
\newblock Distilling free-form natural laws from experimental data.
\newblock \emph{science}, 324\penalty0 (5923):\penalty0 81--85, 2009.

\bibitem[Seeger-Nukpezah et~al.(2015)Seeger-Nukpezah, Geynisman, Nikonova, Benzing, and Golemis]{SeegerNukpezah2015}
Tamina Seeger-Nukpezah, Daniel~M. Geynisman, Anna~S. Nikonova, Thomas Benzing, and Erica~A. Golemis.
\newblock The hallmarks of cancer: relevance to the pathogenesis of polycystic kidney disease.
\newblock \emph{Nature Reviews Nephrology}, 11\penalty0 (9):\penalty0 515–534, April 2015.
\newblock ISSN 1759-507X.
\newblock \doi{10.1038/nrneph.2015.46}.
\newblock URL \url{http://dx.doi.org/10.1038/nrneph.2015.46}.

\bibitem[Seeley(1989)]{seeley1989honey}
Thomas~D Seeley.
\newblock The honey bee colony as a superorganism.
\newblock \emph{American Scientist}, 77\penalty0 (6):\penalty0 546--553, 1989.

\bibitem[Shojaee et~al.(2025)Shojaee, Meidani, Gupta, Farimani, and Reddy]{shojaee2025llmsr}
Parshin Shojaee, Kazem Meidani, Shashank Gupta, Amir~Barati Farimani, and Chandan~K. Reddy.
\newblock {LLM}-{SR}: Scientific equation discovery via programming with large language models.
\newblock In \emph{The Thirteenth International Conference on Learning Representations}, 2025.
\newblock URL \url{https://openreview.net/forum?id=m2nmp8P5in}.

\bibitem[Skuta et~al.(2017)Skuta, Popr, Muller, Jindrich, Kahle, Sedlak, Svozil, and Bartunek]{skuta2017probes}
Ctibor Skuta, Martin Popr, Tomas Muller, Jindrich Jindrich, Michal Kahle, David Sedlak, Daniel Svozil, and Petr Bartunek.
\newblock Probes \& drugs portal: an interactive, open data resource for chemical biology.
\newblock \emph{Nature methods}, 14\penalty0 (8):\penalty0 759--760, 2017.

\bibitem[Spirtes et~al.(2000)Spirtes, Glymour, and Scheines]{spirtes2000causation}
Peter Spirtes, Clark~N Glymour, and Richard Scheines.
\newblock \emph{Causation, prediction, and search}.
\newblock MIT press, 2000.

\bibitem[Tomczak et~al.(2015)Tomczak, Czerwi{\'n}ska, and Wiznerowicz]{tomczak2015review}
Katarzyna Tomczak, Patrycja Czerwi{\'n}ska, and Maciej Wiznerowicz.
\newblock Review the cancer genome atlas (tcga): an immeasurable source of knowledge.
\newblock \emph{Contemporary Oncology/Wsp{\'o}{\l}czesna Onkologia}, 2015\penalty0 (1):\penalty0 68--77, 2015.

\bibitem[Trinh et~al.(2024)Trinh, Wu, Le, He, and Luong]{trinh2024solving}
Trieu~H Trinh, Yuhuai Wu, Quoc~V Le, He~He, and Thang Luong.
\newblock Solving olympiad geometry without human demonstrations.
\newblock \emph{Nature}, 625\penalty0 (7995):\penalty0 476--482, 2024.

\bibitem[Tsherniak et~al.(2017)Tsherniak, Vazquez, Montgomery, Weir, Kryukov, Cowley, Gill, Harrington, Pantel, Krill-Burger, Meyers, Ali, Goodale, Lee, Jiang, Hsiao, Gerath, Howell, Merkel, Ghandi, Garraway, Root, Golub, Boehm, and Hahn]{Tsherniak2017}
Aviad Tsherniak, Francisca Vazquez, Phil~G. Montgomery, Barbara~A. Weir, Gregory Kryukov, Glenn~S. Cowley, Stanley Gill, William~F. Harrington, Sasha Pantel, John~M. Krill-Burger, Robin~M. Meyers, Levi Ali, Amy Goodale, Yenarae Lee, Guozhi Jiang, Jessica Hsiao, William~F.J. Gerath, Sara Howell, Erin Merkel, Mahmoud Ghandi, Levi~A. Garraway, David~E. Root, Todd~R. Golub, Jesse~S. Boehm, and William~C. Hahn.
\newblock Defining a cancer dependency map.
\newblock \emph{Cell}, 170\penalty0 (3):\penalty0 564--576.e16, July 2017.
\newblock ISSN 0092-8674.
\newblock \doi{10.1016/j.cell.2017.06.010}.
\newblock URL \url{http://dx.doi.org/10.1016/j.cell.2017.06.010}.

\bibitem[Udrescu \& Tegmark(2020)Udrescu and Tegmark]{udrescu2020ai}
Silviu-Marian Udrescu and Max Tegmark.
\newblock Ai feynman: A physics-inspired method for symbolic regression.
\newblock \emph{Science advances}, 6\penalty0 (16):\penalty0 eaay2631, 2020.

\bibitem[Wang et~al.(2025)Wang, Skreta, Ser, Gao, Kong, Strieth-Kalthoff, Duan, Zhuang, Yu, Zhu, Du, Aspuru-Guzik, Neklyudov, and Zhang]{wang2025efficient}
Haorui Wang, Marta Skreta, Cher~Tian Ser, Wenhao Gao, Lingkai Kong, Felix Strieth-Kalthoff, Chenru Duan, Yuchen Zhuang, Yue Yu, Yanqiao Zhu, Yuanqi Du, Alan Aspuru-Guzik, Kirill Neklyudov, and Chao Zhang.
\newblock Efficient evolutionary search over chemical space with large language models.
\newblock In \emph{The Thirteenth International Conference on Learning Representations}, 2025.
\newblock URL \url{https://openreview.net/forum?id=awWiNvQwf3}.

\bibitem[Wang et~al.(2021)Wang, Chen, Dang, Zhang, Wang, Yin, Jia, and Zhang]{Wang2021birc}
Jingyuan Wang, Min Chen, Chengxue Dang, Hao Zhang, Xin Wang, Jianhao Yin, Rui Jia, and Yong Zhang.
\newblock The early diagnostic and prognostic value of birc5 in clear-cell renal cell carcinoma based on the cancer genome atlas data.
\newblock \emph{Urologia Internationalis}, 106\penalty0 (4):\penalty0 344–351, July 2021.
\newblock ISSN 1423-0399.
\newblock \doi{10.1159/000517310}.
\newblock URL \url{http://dx.doi.org/10.1159/000517310}.

\bibitem[Weinstein et~al.(2013)Weinstein, Collisson, Mills, Shaw, Ozenberger, Ellrott, Shmulevich, Sander, and Stuart]{weinstein2013cancer}
John~N Weinstein, Eric~A Collisson, Gordon~B Mills, Kenna~R Shaw, Brad~A Ozenberger, Kyle Ellrott, Ilya Shmulevich, Chris Sander, and Joshua~M Stuart.
\newblock The cancer genome atlas pan-cancer analysis project.
\newblock \emph{Nature genetics}, 45\penalty0 (10):\penalty0 1113--1120, 2013.

\bibitem[Weisberg \& Muldoon(2009)Weisberg and Muldoon]{weisberg2009epistemic}
Michael Weisberg and Ryan Muldoon.
\newblock Epistemic landscapes and the division of cognitive labor.
\newblock \emph{Philosophy of science}, 76\penalty0 (2):\penalty0 225--252, 2009.

\bibitem[Wu et~al.(2024)Wu, Bansal, Zhang, Wu, Li, Zhu, Jiang, Zhang, Zhang, Liu, et~al.]{wu2024autogen}
Qingyun Wu, Gagan Bansal, Jieyu Zhang, Yiran Wu, Beibin Li, Erkang Zhu, Li~Jiang, Xiaoyun Zhang, Shaokun Zhang, Jiale Liu, et~al.
\newblock Autogen: Enabling next-gen llm applications via multi-agent conversations.
\newblock In \emph{First Conference on Language Modeling}, 2024.

\bibitem[Xie et~al.(2025)Xie, Zhang, Li, Guo, Guo, Wang, Ren, Yu, and Wu]{Xie2025}
Xin Xie, Yadi Zhang, Yuanyuan Li, Na~Guo, Xiaomeng Guo, Feilong Wang, Kuiwu Ren, Jiangtao Yu, and Fengrui Wu.
\newblock Deubiquitinase <scp>otud4</scp> stabilizes <scp>slc5a2</scp> to promote pancreatic cancer proliferation and migration through enchaining glycolysis‐mediated autophagy.
\newblock \emph{The FASEB Journal}, 39\penalty0 (14), July 2025.
\newblock ISSN 1530-6860.
\newblock \doi{10.1096/fj.202501018r}.
\newblock URL \url{http://dx.doi.org/10.1096/fj.202501018R}.

\bibitem[Yang et~al.(2024)Yang, Chen, Zheng, Zeng, Zhou, Chen, Su, Wang, Wang, Wang, Wang, Jin, Bo, Chen, and Wang]{Yang2024}
Yi~Yang, Bo~Chen, Chongming Zheng, Hao Zeng, Junxi Zhou, Yaqing Chen, Qing Su, Jingxian Wang, Juejin Wang, Yurong Wang, Hongli Wang, Ruxue Jin, Zhiyuan Bo, Gang Chen, and Yi~Wang.
\newblock Association of glucose-lowering drug target and risk of gastrointestinal cancer: a mendelian randomization study.
\newblock \emph{Cell \&amp; Bioscience}, 14\penalty0 (1), March 2024.
\newblock ISSN 2045-3701.
\newblock \doi{10.1186/s13578-024-01214-8}.
\newblock URL \url{http://dx.doi.org/10.1186/s13578-024-01214-8}.

\bibitem[Yao et~al.(2023)Yao, Zhao, Yu, Du, Shafran, Narasimhan, and Cao]{yao2023react}
Shunyu Yao, Jeffrey Zhao, Dian Yu, Nan Du, Izhak Shafran, Karthik Narasimhan, and Yuan Cao.
\newblock React: Synergizing reasoning and acting in language models.
\newblock In \emph{International Conference on Learning Representations (ICLR)}, 2023.

\bibitem[Zheng et~al.(2018)Zheng, Aragam, Ravikumar, and Xing]{zheng2018dags}
Xun Zheng, Bryon Aragam, Pradeep~K Ravikumar, and Eric~P Xing.
\newblock Dags with no tears: Continuous optimization for structure learning.
\newblock \emph{Advances in neural information processing systems}, 31, 2018.

\bibitem[Zheng et~al.(2025)Zheng, Jiang, Qi, and Peng]{Zheng2025}
Yuxiao Zheng, Lei Jiang, Feng Qi, and Bo~Peng.
\newblock Cop1 drives renal cell carcinoma progression by targeting acsl4 for ubiquitin-mediated degradation and inhibiting ferroptosis.
\newblock \emph{Frontiers in Oncology}, 15, May 2025.
\newblock ISSN 2234-943X.
\newblock \doi{10.3389/fonc.2025.1570727}.
\newblock URL \url{http://dx.doi.org/10.3389/fonc.2025.1570727}.

\bibitem[Zhuge et~al.(2023)Zhuge, Liu, Faccio, Ashley, Csord{\'a}s, Gopalakrishnan, Hamdi, Hammoud, Herrmann, Irie, et~al.]{zhuge2023mindstorms}
Mingchen Zhuge, Haozhe Liu, Francesco Faccio, Dylan~R Ashley, R{\'o}bert Csord{\'a}s, Anand Gopalakrishnan, Abdullah Hamdi, Hasan Abed Al~Kader Hammoud, Vincent Herrmann, Kazuki Irie, et~al.
\newblock Mindstorms in natural language-based societies of mind.
\newblock \emph{arXiv preprint arXiv:2305.17066}, 2023.

\bibitem[Zou et~al.(2019)Zou, Palte, Deik, Li, Eaton, Wang, Tseng, Deasy, Kost-Alimova, Dančík, Leshchiner, Viswanathan, Signoretti, Choueiri, Boehm, Wagner, Doench, Clish, Clemons, and Schreiber]{Zou2019gpx4}
Yilong Zou, Michael~J. Palte, Amy~A. Deik, Haoxin Li, John~K. Eaton, Wenyu Wang, Yuen-Yi Tseng, Rebecca Deasy, Maria Kost-Alimova, Vlado Dančík, Elizaveta~S. Leshchiner, Vasanthi~S. Viswanathan, Sabina Signoretti, Toni~K. Choueiri, Jesse~S. Boehm, Bridget~K. Wagner, John~G. Doench, Clary~B. Clish, Paul~A. Clemons, and Stuart~L. Schreiber.
\newblock A gpx4-dependent cancer cell state underlies the clear-cell morphology and confers sensitivity to ferroptosis.
\newblock \emph{Nature Communications}, 10\penalty0 (1), April 2019.
\newblock ISSN 2041-1723.
\newblock \doi{10.1038/s41467-019-09277-9}.
\newblock URL \url{http://dx.doi.org/10.1038/s41467-019-09277-9}.

\end{thebibliography}
\bibliographystyle{iclr2026_conference}

\newpage

\appendix

\section{Extended Related Works}
\label{app:extended_rws}

Our work integrates over several prior directions, which we detail below.

\textbf{Data-driven discovery.}~Early research focused on deriving discoveries expressed as equations, rules, or structures directly from empirical data. Fields such as \textit{symbolic regression} recover closed-form mathematical equations from measurements \citep{schmidt2009distilling, brunton2016discovering, udrescu2020ai}, while logic programming and rule discovery uncover hypotheses expressed as relational or \textit{propositional rules} in discrete domains \citep{quinlan1990learning, clark1989cn2, lin2020generalized}. A related thread is causal discovery, which seeks to infer underlying \textit{causal graphs} from observational data using independence constraints, scoring criteria, or functional assumptions \citep{spirtes2000causation, zheng2018dags, peters2014causal}. 

\textbf{LLM-augmented discovery.}~Recent work have investigated replacing ad-hoc inductive biases with the scientific priors encoded in LLMs. Here, LLMs are employed in specialized roles, as \textbf{search operators} to generate and modify hypotheses (commonly expressed in code), guided by formal evaluators (e.g., solvers, experiments, or reward signals) providing feedback. This framework has enabled the discovery of new algorithms and mathematical constructs \citep{romera2024mathematical, novikov2025alphaevolve}, and has been applied across domains including neural architecture search \citep{chen2023evoprompting}, interpretable decision trees \citep{liu2025decision}, symbolic equations \citep{shojaee2025llmsr}, formal theorems \citep{trinh2024solving}, robotics reward functions \citep{ma2024eureka}, and molecular design \citep{wang2025efficient}. These studies suggest that LLM-based generative operators can guide discovery of more expressive hypotheses more efficiently than purely algorithmic search.

\textbf{Agentic science.}~An emerging theme considers \textit{agentic} AI systems that combine LLMs with external tools and memory to automate different aspects of the scientific process. One line of work emphasizes \textbf{automation of experimental workflows}, focusing on the orchestration and execution of experiments—from planning chemical synthesis or biomedical analyses to coordinating CRISPR-based pipelines \citep{m2024augmenting, ruan2024automatic, huang2025biomni, qu2025crispr}. Distinct from this, and more directly relevant to our work, is research on \textbf{hypothesis generation and refinement}, where LLM-based agents autonomously propose, critique, and evolve scientific ideas. Seminal examples include the \texttt{AI Scientist} \citep{lu2024ai}, which is able to generate research ideas, write code, run experiments, analyze results, and draft complete research papers; and the \texttt{AI Co-Scientist} \citep{gottweis2025towards}, a multi-agent system that employs a “generate–debate–evolve” cycle to formulate and refine hypotheses, particularly in biomedical domains. Also related is work on hypothesis falsification, where agents conduct sequential hypothesis testing under rigorous statistical control \citep{huang2025automated}, though this line of research focuses exclusively on falsification. Similar projects (e.g.\citet{saeedi2025astroagents,ghafarollahi2025sciagents}) illustrate domain-tailored instantiations of this paradigm.

\textbf{Distributed systems.} Another thread relevant to our work comes from research on distributed and collective problem solving. Classical \emph{swarm intelligence} algorithms, such as Ant Colony Optimization \citep{dorigo1997ant}, Particle Swarm Optimization \citep{kennedy1995particle}, and Bee Colony models \citep{seeley1989honey}, demonstrate how simple interacting agents can collectively explore large search spaces more effectively than any single agent. Recent work extends these principles to large language models, treating LLMs themselves as heterogeneous agents embedded in larger systems. \texttt{Generative Agents} \citep{park2023generative} simulate human-like social interactions with memory and reflection, while recent works have extended this to large-scale agent-based simulations with LLM agents \citep{zhuge2023mindstorms,gao2024large}. These approaches echo longstanding ideas such as Minsky’s \emph{Society of Mind} \citep{minsky1986society}, where cognition arises from the interaction of specialized but simple agents, and motivate the design of agentic scientific systems that integrate memory, specialization, and collective or emergent behavior.

\clearpage
\section{Additional Technical Details}
\label{app:additional_details}

\subsection{Registry and Archive Schema}
\label{app:schema}

To support persistent storage and retrieval of information in \texttt{ASCollab}, we define schemas for both the \textbf{agent registry} and the \textbf{internal archive}. The registry maintains structured profiles of each agent in the system, while the archive stores metadata about submitted manuscripts, including review information and bibliographic links. Together, these schemas enable reproducibility, traceability, and analysis of the evolving research ecosystem.

\Cref{lst:papermetadata} shows the \texttt{PaperMetadata} dataclass, which records all key information about a manuscript submitted to the archive. This includes authorship (the primary agent and collaborators), bibliographic attributes (title, abstract, manuscript text), impact measures (citation counts), temporal information (publication time), and optional artifacts such as executable code. The \texttt{cited\_paper\_ids} field enables linking between papers in the archive, while the \texttt{metareview} field stores evaluation results when available.

\begin{lstlisting}[style=pyclean,caption={Schema for paper metadata entries in the internal archive.},label={lst:papermetadata}]
@dataclass
class PaperMetadata:
    paper_id: str
    primary_agent_id: str
    collab_agent_ids: List[str]
    title: str
    abstract: str
    manuscript: str
    citation_count: int 
    publication_t: int
    cited_paper_ids: List[Dict[str, str]]
    code_script: Optional[str] = None
    metareview: Optional[PaperMetaReview] = None
    status: str
\end{lstlisting}

Reviews are represented using the \texttt{PaperMetaReview} dataclass (\Cref{lst:papermetareview}). Each metareview corresponds to one paper and captures textual justification, a numeric score, ranking, and the final decision outcome. This allows the archive to track not only papers but also the evaluation criteria applied to them.

\begin{lstlisting}[style=pyclean,caption={Schema for metareview entries associated with submitted papers.},label={lst:papermetareview}]
@dataclass
class PaperMetaReview:
    paper_id: str
    meta_review_text: str
    overall_score: float
    rank: str
    justification: str
    decision: str
\end{lstlisting}

Finally, the agent registry maintains structured information about each research agent through the \texttt{AgentProfile} dataclass (\Cref{lst:agentprofile}). These profiles capture identifiers, epistemic behavior, and domain expertise, along with performance metrics such as citation counts and the number of accepted papers. This registry is essential for analyzing heterogeneity and longitudinal contributions of agents in the system.

\begin{lstlisting}[style=pyclean,caption={Schema for agent profile entries in the registry.},label={lst:agentprofile}]
@dataclass
class AgentProfile:
    agent_id: str
    behavior: str
    expertise: str
    expertise_topics: List[str]
    citation_count: int
    num_accepted_papers: int
\end{lstlisting}

\subsection{Scientific Personas}
\label{app:epistemic_profiles}

To introduce structured heterogeneity into the agent population, we prompt the underlying LLM to generate distinct \emph{scientific personas}. Each persona reflects a unique epistemic stance and domain expertise, ensuring diversity in how agents approach idea generation, collaboration, scope, evaluation, literature use, and resource allocation. We define two schema templates that guide the generation of these personas: one for epistemic behavior and one for technical expertise. 

\Cref{lst:persona_epistemic} shows the schema used to elicit \textbf{epistemic researcher profiles}. In addition to epistemic orientation, each agent is assigned a \textbf{domain expertise profile}, defined with respect to specific datasets and methodological skills. The schema in \Cref{lst:persona_expertise} ensures that expertise is expressed as concrete, methodological capabilities (e.g., statistical models, validation strategies, pitfalls).

\begin{lstlisting}[style=prompt,caption={Schema for epistemic researcher personas generated at system initialization.},label={lst:persona_epistemic}]
You are to generate a single epistemic researcher profile.

The profile should:
- Be written in second person (e.g., ``You are'').
- Be returned in bullet point form (one bullet per stance).
- Contain exactly one distinct persona per completion.

Each persona must capture how the researcher behaves and thinks across six stances:
1. Ideas - Refining and extending existing ideas <-> generating brand new ones.
2. Collaboration - Independence <-> collaboration.
3. Scope - Broad exploration <-> deep exploitation of a problem.
4. Evaluation - Critical scrutiny <-> constructive engagement.
5. Literature - Reliance on existing literature <-> intuition with minimal reference to prior work.
6. Resources - Maximal use of resources and depth <-> lean, minimalist approaches.

Requirements:
- Generate exactly one persona per completion.
- Provide exactly six bullet points, one for each stance.
- Each bullet point must begin with "When it comes to [stance]:" followed by the persona's orientation.
- Keep each bullet concise, vivid, and natural-sounding.
- The persona should reflect an expert researcher with a unique epistemic orientation and personality.
- Return only the bullet point profile, with no labels, numbers, or extra commentary.
\end{lstlisting}

\begin{lstlisting}[style=prompt,caption={Schema for domain expertise profiles describing technical methods and approaches.},label={lst:persona_expertise}]
You are to generate a domain expertise description for a researcher with the following specific technical expertise areas: {topics_str}.

{dataset_context}

The expertise should describe what domain knowledge and technical skills this researcher possesses in these areas, specifically focused on how they would generate novel research findings using the available datasets. Focus on concrete methods, approaches, and practical knowledge for conducting innovative research rather than generic descriptions.

IMPORTANT: The expertise should be pan-cancer and generalized - describe technical methods and computational approaches that can be applied broadly across different cancer types and biological contexts, rather than being specific to any particular cancer type (e.g., kidney cancer, breast cancer, etc.). Focus on the methodological and technical aspects that would lead to novel discoveries when working with these specific datasets to generate breakthrough research findings.

Output Requirements:
- Generate exactly one bullet point for each of the {len(selected_topics)} topics provided, in the same order.
- Each bullet point must be written in second person ("You...") and describe specific technical skills/knowledge for generating novel findings.
- Keep each bullet to 1-2 sentences.
- Be specific about methods, models, metrics, pitfalls, validation strategies, or practical considerations for research discovery.
- Focus on how the researcher would use these skills to generate new insights from the available datasets.
- Avoid generic phrases like "data science" or "machine learning" without specific qualifiers.
- Avoid references to specific cancer types - keep descriptions general and broadly applicable.
- No labels, numbers, or extra commentary outside the bullets.

Format your response as:
<expertise>
- You ...
- You ...
- You ...
</expertise>
\end{lstlisting}

\subsection{Final Report, Review, and Metareview Instructions}
\label{app:agent_prompts}

Each agent is given explicit output instructions to ensure that generated reports, reviews, and meta-reviews follow a consistent structure. These schemas serve both as constraints and as templates for evaluation, making it possible to systematically compare and archive agent contributions. We define three main instruction sets: (i) \emph{Final Report Requirements}, (ii) \emph{Evaluation Criteria for Reviews}, and (iii) \emph{Meta-Review Structure}. 

\Cref{lst:final_report} specifies the structure of the \textbf{Final Report}, which every research agent must prepare before exhausting its budget. The schema enforces a set of mandatory sections (e.g., title, hypothesis, evidence, limitations, references), and emphasizes the use of properly retrieved citations.

\begin{lstlisting}[style=prompt,caption={Schema for agent Final Report output, including mandatory sections and formatting requirements.},label={lst:final_report}]
When you feel ready, prepare a concise, clear, and well-structured Final Report (you must do so before running out of budget) with the following sections:

Final Report structure (mandatory sections):
# Title
(A concise, representative title of your findings.)

# Research Question
(A single, clear question your hypothesis addresses.)

# Hypothesis and Key Findings
(A concise statement of your hypothesis and the main findings that support it.)

# Rationale/Mechanism
(Brief explanation of why this finding makes sense.)

# Empirical Evidence
(Bullet list of dataset findings supporting the finding. Include metrics, statistical tests, graphs, or model outputs, synthesized and not just raw dumps. Include relevant details on analysis methods.)

# Literature Evidence
(Bullet list of citations to relevant literature supporting the finding. Include brief summaries of key findings from each paper and how they relate to the hypothesis. Your finding should be novel and not just a repeat of prior work, but prior work can provide supporting context.)

# Assumptions
(Explicitly list assumptions that underlie the hypothesis.)

# Limitations
(Explicitly list possible caveats or alternative explanations)

# References
List of cited papers with full citations in a consistent format. If you are referencing sources from the open internet, use the following format: 
- Author(s). (Year). Title of the article. Title of the Journal, Volume(Issue), page range (if applicable). 
If you are referencing sources from the internal paper archive, please use the following format: 
- [Internal Archive] {'paper_id': <paper_id>, 'agent_id': <agent_id>, 'title': <title>}

Instructions:
- Use only retrieved references; do not fabricate citations.
- List all references in a References section using the formats below:
    Internal paper archive:
    - [Internal Archive] {'paper_id': <paper_id>, 'agent_id': <agent_id>, 'title': <title>}
    External sources:
    - Author(s). (Year). Title of the article. Title of the Journal, Volume(Issue), page range.
\end{lstlisting}

To evaluate submitted reports, reviewer agents are prompted with the schema in \Cref{lst:evaluation}, which covers both qualitative criteria (summary, motivation, claims, methodology, novelty, significance) and quantitative ratings (support, soundness, significance, originality, overall recommendation). This ensures that each review is structured, comparable, and comprehensive.

\begin{lstlisting}[style=prompt,caption={Schema for reviewer evaluation criteria and quantitative rating scales.},label={lst:evaluation}]
Evaluation Criteria:
1. Summary:
Briefly summarize the report (including the main findings, main results, etc. that the report claims to contribute). This summary should be objective, and not be used to critique the report. A well-written summary should not be disputed by the authors of the report or other readers.

2. Motivation:
- What is the specific question and/or problem tackled by the report?
- Is the problem well motivated and clearly situated in the broader literature?

3. Claims and Evidence:
- Are the main claims of the report clearly stated? Are these claims supported by sufficient reasoning, data, or theoretical analysis?
- If evidence is lacking, which claims are problematic and why?

4. Soundness of Methodology:
- Are the methods and/or analyses and/or evaluation metrics appropriate for the problem?
- Are the designs, assumptions, and evaluation criteria scientifically valid?
- NOTE: you do not have to reproduce the results (i.e., run the code, etc), but you should evaluate whether the methodology is sound and appropriate.

5. Relation to Prior Knowledge:
- How are the key contributions of the report related to the broader scientific literature? Be specific in terms of prior related findings/results/ideas/etc.
- Do the main findings either extend, challenge, or refine prior work in the field? If so, how?

6. Novelty and Significance:
- What is the significance of the work? Does it contribute new knowledge and sufficient value to the community? 
Are the contributions genuinely new, incremental extensions of prior work, or simply restatements of existing knowledge?
- What is the potential impact or value to the field (empirical, theoretical, practical)?

7. Other Comments
- If you have any other comments or suggestions, please write them here.

# Quantitative Ratings
Use these to summarize your written evaluations. Respond with an integer for each category.

- Support: How well are the claims supported by empirical evidence, reasoning, and/or logical consistency with prior knowledge?
4 = Excellent | 3 = Good | 2 = Fair | 1 = Poor

- Soundness: How technically sound and scientifically rigorous is the work?
4 = Excellent | 3 = Good | 2 = Fair | 1 = Poor

- Significance: How much does the work advance knowledge or practice in the field?
4 = Excellent | 3 = Good | 2 = Fair | 1 = Poor

- Originality: How novel are the ideas, methods, or results?
4 = Excellent | 3 = Good | 2 = Fair | 1 = Poor

- Overall Recommendation:
5: Strong accept
4: Accept
3: Weak accept (i.e., leaning towards accept, but could also be rejected)
2: Weak reject (i.e., leaning towards reject, but could also be accepted)
1: Reject
\end{lstlisting}

Finally, the schema in \Cref{lst:metareview} guides meta-review agents, which synthesize individual reviews and provide a comparative assessment across multiple reports. The template enforces a three-part structure: a brief summary, a comparative analysis, and a final decision including a score, rank, and justification.

\begin{lstlisting}[style=prompt,caption={Schema for meta-review output structure, including summary, comparative analysis, and decision.},label={lst:metareview}]
For each report, provide a meta-review following this exact structure:

Paper ID: <id of the report>

1. Brief Summary
- A 1-2 sentence bullet-point summary of its main contributions.
- A 1-2 sentence bullet-point summary of its strengths and weaknesses, based on the your own judgement and the reviews.

2. Comparative Analysis
- 2-3 bullet points assessing the submission against the criteria.
- Where possible, contrast with other reports (e.g., "significantly more/less novel than report X").

3. Decision
- score: <float between 0 and 1> (assign each report a score on a 0-1 scale, where 1 = best overall quality)
- rank: <integer rank, 1 is best> (assign each submission a rank from 1 to N, where 1 = best. No ties allowed)
- justification: <brief justification> (1-2 sentences for each report's relative position. This should be self-contained and complete without references to other reports)

\end{lstlisting}

\subsection{Agentic Reasoning and Tool-Use}
\label{app:reasoning_tool_use}

Agents in \texttt{ASCollab} reason and act using the \emph{ReAct} paradigm \citep{yao2023react}, which interleaves natural language reasoning with tool invocations. This allows agents to plan, reflect, and take actions in a single loop, enabling both exploratory reasoning and structured data analysis. An agent generates a reasoning trace (“Thought”), selects a tool (“Action”), and integrates the result into its ongoing chain of reasoning. \Cref{lst:react_example} shows a simplified illustration of this reasoning–acting loop.

\begin{lstlisting}[style=prompt,caption={Example of an agent using ReAct-style reasoning to query PubMed and refine a hypothesis.},label={lst:react_example}]
Thought: I want to check whether mutations in KRAS are frequently associated with pancreatic cancer.
Action: PubMed("KRAS pancreatic cancer mutations frequency")
Observation: The retrieved abstracts indicate KRAS mutations occur in >90% of pancreatic ductal adenocarcinomas.
Thought: This supports my hypothesis that KRAS status should be included as a covariate in survival analysis.
\end{lstlisting}

Beyond reasoning, agents have access to a set of scientific software libraries and programmatic tools. These resources enable them to execute analyses spanning differential expression, pathway enrichment, survival modeling, and network inference. The available Python packages are summarized in \Cref{lst:python_packages}, which defines a schema mapping each package to its primary function in transcriptomic, proteomic, or clinical workflows.

\begin{lstlisting}[caption={Schema of Python packages available to agents for omics, pathway, and survival analysis.},label={lst:python_packages}]
{
  "pydeseq2": "Differential expression analysis for bulk RNA-seq (Python reimplementation of DESeq2).",
  "rpy2": "Bridge to R, lets you use DESeq2, edgeR, limma, survival, and other Bioconductor packages from Python.",
  "statsmodels": "Statistical modeling (linear/GLM/mixed models; also duration/survival models) for DE and covariate analysis.",
  "scanpy": "Gene-expression toolkit (QC, normalization, clustering, visualization); can handle bulk matrices via AnnData.",
  "anndata": "Annotated matrix container for expression data with sample/gene metadata, backbone for many omics workflows.",
  "gseapy": "Gene set enrichment (GSEA/Preranked/Enrichr/MSigDB) for pathways from RNA/proteomics gene lists.",
  "gprofiler-official": "g:Profiler client for GO/KEGG/Reactome enrichment and ID conversion.",
  "mygene": "Fast gene ID mapping and annotation (symbols <-> Ensembl/Entrez) for building bulk/proteomics panels.",
  "biomart": "Access Ensembl BioMart to retrieve gene/transcript/protein annotations and mappings.",
  "bioservices": "Programmatic access to bio databases (e.g., UniProt, KEGG, Reactome, ChEMBL) for protein/drug/pathway metadata.",
  "biopython": "General bioinformatics utilities-sequence I/O, Entrez/UniProt access, useful for proteomics ID work.",
  "igraph": "Graph algorithms for pathway/network analysis (centrality, community detection) on gene-protein networks.",
  "networkx": "Network analysis and visualization for pathways/PPIs/drug-target graphs.",
  "leidenalg": "Leiden community detection, useful for clustering genes/proteins in co-expression or PPI networks.",
  "lifelines": "Survival analysis (Kaplan-Meier, Cox PH, AFT, competing risks) for clinical/time-to-event data.",
  "scikit-learn": "Machine learning (feature selection, classification/regression, clustering) for expression/proteomics models.",
  "scikit-bio": "Bioinformatics stats and distances (diversity, ordination); can support multi-omics workflows.",
  "PubChemPy": "Client for PubChem to fetch compound properties, synonyms, assays, handy for drug annotation.",
  "pandas": "Tabular data wrangling, joins/reshapes/IO for expression matrices, proteomics tables, and survival covariates.",
  "numpy": "Numerical arrays and linear algebra underpinning most computations in RNA/proteomics analyses.",
  "openpyxl": "Read/write Excel files, useful for proteomics exports (e.g., MaxQuant/PD) and metadata sheets."
}
\end{lstlisting}

In addition to Python packages, agents can call higher-level tools that enable them to search literature, discover collaborators, and communicate within the agent network. These tools are listed below:

\begin{enumerate}[leftmargin=*,itemsep=0.4ex,parsep=-0.1ex]
    \item \textbf{PubMed:} Wrapper around PubMed for querying biomedical abstracts and literature.
    \item \textbf{SemanticScholar:} Search Semantic Scholar with free-text queries and return summaries.
    \item \textbf{InternalArchive:} Search internally published research papers by topic, methodology, or research area.
    \item \textbf{SearchRegistry:} Retrieve researcher profiles (expertise, citations, papers) from the registry.
    \item \textbf{EstablishCollaboration:} Create a collaboration connection with another researcher by agent ID.
    \item \textbf{Communicate:} Send messages or data payloads to a collaborator, addressing them directly in first person.
\end{enumerate}

\subsection{Human Expert Evaluation}
\label{app:evaluation_rubric}
For the KIRC dataset, we engaged an  domain expert (computational drug discovery with prior KIRC research experience) to score each paper’s central hypothesis. Because evaluation criteria vary and no single standard exists, we adopted two broadly accepted dimensions: \emph{Novelty} (“has this been done before?”) and \emph{Quality} (“does it make sense given prior literature, and is there external corroboration?”).

Each hypothesis was scored on a 1–5 scale for both dimensions using the rubric in Table~\ref{tab:human-eval-rubric}. To reduce subjectivity and bias, the evaluator followed predefined anchors, and applied the same procedure across all items. The evaluator had full access to all run artifacts produced in our experiments as well as to publicly available online resources.

\begin{table*}[t]
\centering
\footnotesize
\caption{\textbf{Human evaluation rubric for novelty (N) and quality (Q).}}
\label{tab:human-eval-rubric}
\setlength{\tabcolsep}{8pt}
\begin{tabular}{c l}
\toprule
\textbf{Dim.} & \textbf{Score} \\
\midrule
\multicolumn{2}{l}{\emph{Novelty (N)}} \\
\midrule
N1 & Already published in essentially the same form \\
N2 & Very similar result published via different methodology \\
N3 & Significant overlap with prior themes/pathways \\
N4 & Minor overlap; clearly new angle or combination \\
N5 & Substantive novel contribution \\
\midrule
\multicolumn{2}{l}{\emph{Quality (Q)}} \\
\midrule
Q1 & Conflicts with strong prior evidence; likely invalid \\
Q2 & Weak/ambiguous support \\
Q3 & Corroborated on the \emph{same} dataset \\
Q4 & Corroborated on \emph{different} dataset/domain \\
Q5 & Strong external validation or literature evidence leading to plausibility \\
\bottomrule
\end{tabular}
\end{table*}

\clearpage
\section{Dataset Details}
\label{app:dataset_details}

We analyze three TCGA cohorts—\textbf{PAAD} (Pancreatic Adenocarcinoma) \citep{raphael2017integrated}, \textbf{KIRC} (Kidney Renal Clear Cell Carcinoma) \citep{cancer2013comprehensive}, and \textbf{DLBC} (Diffuse Large B-Cell Lymphoma) \citep{weinstein2013cancer}—using matched multi-omics resources where available. Unless stated otherwise, bulk RNA-seq matrices are Illumina HiSeq (polyA\,+) with gene-level \(\log_2(x{+}1)\) RSEM-normalized counts mapped via UCSC Xena HUGO probeMap; RPPA is the TCGA reverse-phase protein array panel (normalized intensities); PARADIGM IPL provides integrated pathway levels derived from RNA-seq and copy-number within a curated interaction graph; and survival files contain overall- and disease-specific survival endpoints. TCGA barcodes follow the standard suffix convention (“\(-01\)” tumour, “\(-11\)” solid-tissue normal). For cross-modal analyses we restrict to the intersection of barcodes shared by the relevant matrices.

\paragraph{Summary of sample counts.}
Table \ref{tab:dataset-sample-counts} lists the number of samples per cohort and modality used in this study. 

\begin{table}[h]
\centering
\scriptsize
\begin{tabular}{lcccc}
\toprule
\textbf{Cohort} & \textbf{Bulk RNA-seq (samples)} & \textbf{RPPA (samples)} & \textbf{PARADIGM IPL (samples)} & \textbf{Survival (rows)} \\
\midrule
PAAD & 183 & 123 & 176 & 196\\
KIRC & 606 & 478 & 507 & 944 \\
DLBC & 48 &33 & 48 & 48 \\
\bottomrule
\end{tabular}
\caption{Sample counts per modality for TCGA PAAD, KIRC, and DLBC.}
\label{tab:dataset-sample-counts}
\end{table}

\subsection*{Per-modality descriptions (shared across cohorts)}

\paragraph{Bulk RNA-seq (polyA\,+ Illumina HiSeq).}
Gene-level expression matrices are provided as \(\log_2(x{+}1)\) RSEM-normalized counts with UCSC Xena HUGO gene identifiers (rows = genes, columns = samples). We use tumour/normal splits via barcode suffixes (“01” vs.\ “11”) and, when combining with survival, subset to overlapping barcodes. No re-normalization or batch correction is applied unless explicitly noted in the experiment section.

\paragraph{RPPA (Reverse-Phase Protein Array).}
RPPA assays quantify total and modified protein features using antibody-based arrays (rows = protein features, columns = samples). We use TCGA-normalized values as distributed. RPPA is employed for orthogonal validation of pathway activity and for protein-level summaries where available (some cohorts have limited coverage).

\paragraph{PARADIGM Integrated Pathway Levels (IPL).}
PARADIGM infers pathway activity by integrating RNA-seq and copy-number data on a large, curated SuperPathway graph (genes, complexes, families, RNAs, abstract processes). The resulting matrix (rows = pathway features; columns = samples) provides pathway-level readouts complementary to gene-level expression. We use the distributed IPL values without additional scaling.

\paragraph{Clinical survival.}
The survival table contains overall survival (\texttt{OS}, event indicator) and times in days (\texttt{OS.time}, \texttt{DSS.time} where available). Row indices are TCGA barcodes. Agents can combine molecular and survival data for more comprehensive analysis.

\clearpage
\section{Agentic Case Studies: Rediscovery, Extension, and Novel Proposals}
\label{app:agentic_case_studies}
We illustrate the capabilities of our agentic system through concise case studies and links to prior work—an approach that is more informative than aggregate metrics given the inherent difficulty of hypothesis evaluation. To keep the setting realistic, all case studies are drawn from the top 50 highest-rated accepted papers. In three representative examples, the agents (i) independently rediscover key analyses, (ii) extend prior findings with additional evidence, and (iii) propose mechanistic hypotheses that we validate using DepMap \citep{Tsherniak2017}. The reports have been typeset for clarity; all content remains unchanged.

\paragraph{Negative cases (rejections).}
Beyond positive results, we include counterexamples where our review pipeline recommends rejection. These illustrate how the system identifies overlap with established literature, flags inadequate support or implausible mechanisms, and aligns its decisions with documented prior evidence. Together, the positive and negative cases clarify both the strengths and the boundaries of the agentic approach.

\subsection{Case Study 1: Role of ACSL4, GPX4, and FTH1 in KIRC}
\label{case_study_1}
This report (\emph{Expanding Ferroptosis-Targeting Strategies in Kidney Renal Clear Cell Carcinoma (KIRC): Therapeutic Potential of ACSL4, GPX4, and FTH1}) builds directly on prior agent work (\emph{Targeting Ferroptosis Pathways via SLC7A11 and ALOX5 Inhibitors for Therapeutic Intervention in KIRC}) while extending the ferroptosis axis beyond SLC7A11/ALOX5 to ACSL4, GPX4, and FTH1. Prior literature had noted gaps and mixed evidence: the expression and prognostic value of \emph{ACSL4} in ccRCC remained incompletely understood \citep{Guo2015}; \emph{FTH1} had been reported as differentially expressed in isolation \citep{Huang2019fth1}; and \emph{GPX4} had likewise been highlighted independently \citep{Zou2019gpx4}. External functional data from DepMap further support target plausibility, showing \emph{significantly reduced proliferation upon gene knockout} (CHRONOS scores: \emph{FTH1} $-0.7432$, A mechanistic link between ubiquitin signaling and ferroptosis in RCC via ACSL4 is suggested by the study titled \emph{“COP1 drives renal cell carcinoma progression by targeting ACSL4 for ubiquitin-mediated degradation and inhibiting ferroptosis”} published in May~2025 \citep{Zheng2025}—after the GPT-4o knowledge cutoff—and, importantly, neither agent surfaced or queried that paper during generation. A separate November~2024 work proposes a different role for \emph{ACSL4} (post-cutoff for our baseline system). We additionally note that our model posits a slightly different role for \emph{ACSL4} within the ferroptosis pathway relative to earlier agent analyses. To verify novelty and positioning, we systematically searched PubMed and Google for these genes in relation to kidney cancer; to the best of our knowledge and as reported by the authors, this is the first report integrating \emph{ACSL4}, \emph{GPX4}, and \emph{FTH1} together in the KIRC ferroptosis context while cross-referencing DepMap essentiality and co-targeting considerations.

\definecolor{lstbg}{RGB}{248,248,248}
\definecolor{lstframe}{RGB}{220,220,220}
\lstdefinestyle{pyclean}{
  language=Python,
  basicstyle=\ttfamily\small,
  numbers=left,
  numberstyle=\scriptsize\color{gray},
  stepnumber=1,
  numbersep=8pt,
  showstringspaces=false,
  breaklines=true,
  frame=single,
  framerule=0.5pt,
  rulecolor=\color{lstframe},
  backgroundcolor=\color{lstbg},
  tabsize=2,
  keepspaces=true
}

\subsubsection*{Expanding Ferroptosis-Targeting Strategies in KIRC: ACSL4, GPX4, and FTH1}
\label{sec:ferroptosis-kirc}

\paragraph{Research Question.}
Can ACSL4, GPX4, and FTH1 serve as therapeutic targets in KIRC, and can their inhibitors enhance ferroptosis pathways for improved patient outcomes?

\paragraph{Hypothesis and Key Findings.}
ACSL4, GPX4, and FTH1—key regulators of ferroptosis—are significantly dysregulated in KIRC and associated with survival outcomes. Therapeutic modulation using inhibitors such as \emph{troglitazone} (ACSL4), \emph{RSL3} (GPX4), and \emph{PD194953} (FTH1) offers promising strategies to enhance ferroptosis and suppress tumor growth.

\paragraph{Rationale / Mechanism.}
Ferroptosis is an iron-dependent form of regulated cell death driven by lipid peroxidation. Dysregulation of ferroptosis-related genes enables tumor cells to evade cell death and promotes progression. Targeting ACSL4 (long-chain fatty acid metabolism), GPX4 (antioxidant defense), and FTH1 (iron storage) can disrupt tumor survival mechanisms and induce ferroptosis.

\begin{table}[t]
\centering
\small
\setlength{\tabcolsep}{8pt}
\caption{\textbf{Differential expression of ferroptosis genes in KIRC.} Negative log2 fold change (LFC) indicates downregulation in tumors.}
\label{tab:ferroptosis-kirc-de}
\begin{tabular}{lccc}
\toprule
\textbf{Gene} & \textbf{Direction in Tumor} & \textbf{Log2 FC} & \textbf{$p$-value} \\
\midrule
ACSL4 & Downregulated & $-1.51$ & $3.48\times 10^{-37}$ \\
GPX4  & Upregulated   & $\,\,0.47$ & $1.99\times 10^{-9}$ \\
FTH1  & Upregulated   & $\,\,0.89$ & $7.98\times 10^{-23}$ \\
\bottomrule
\end{tabular}
\end{table}

\paragraph{Empirical Evidence.}
\begin{itemize}
  \item \textbf{Differential expression:} See Table~\ref{tab:ferroptosis-kirc-de}.
  \item \textbf{Survival correlations:} ACSL4, GPX4, and FTH1 exhibit significant correlations with survival outcomes ($p<0.05$).
  \item \textbf{Pathway enrichment:} Enrichment in \emph{Ferroptosis} ($p=2.94\times 10^{-12}$) and \emph{Regulation of ferroptosis} ($p=4.96\times 10^{-6}$).
\end{itemize}

\begin{table}[t]
\centering
\small
\setlength{\tabcolsep}{6pt}
\caption{\textbf{Candidate inhibitors targeting ferroptosis regulators.}}
\label{tab:ferroptosis-kirc-drugs}
\begin{tabular}{lll}
\toprule
\textbf{Target} & \textbf{Inhibitor} & \textbf{Note} \\
\midrule
ACSL4 & Troglitazone & Modulates lipid acylation \\
GPX4  & RSL3         & Covalent GPX4 inhibitor; induces ferroptosis \\
FTH1  & PD194953     & Iron storage modulation \\
\bottomrule
\end{tabular}
\end{table}

\paragraph{Assumptions.}
\begin{itemize}
  \item Dysregulation of ACSL4, GPX4, and FTH1 contributes to KIRC progression via ferroptosis inhibition.
  \item The listed inhibitors specifically and effectively modulate the intended targets in KIRC.
\end{itemize}

\paragraph{Limitations.}
\begin{itemize}
  \item Protein-level validation of ACSL4, GPX4, and FTH1 in KIRC is currently unavailable.
  \item KIRC-specific experimental validation of inhibitor efficacy remains to be performed.
\end{itemize}

\paragraph{Literature and Prior Evidence.}
\begin{itemize}
  \item Internal Archive: \textit{Title:} \emph{Targeting Ferroptosis Pathways via SLC7A11 and ALOX5 Inhibitors for Therapeutic Intervention in Kidney Renal Clear Cell Carcinoma (KIRC)}
  \item Internal Archive: \textit{Title:} \emph{Targeting Ferroptosis Pathways in Kidney Renal Clear Cell Carcinoma: Therapeutic Implications of SLC7A11 and NCOA4}
  \item PubMed: Chrysin enhances sunitinib sensitivity in renal cell carcinoma by inducing ferroptosis via targeting PI3K/Akt/GPX4 pathway. Elsevier, 2025.
  \item PubMed: tRNA-derived small RNAs: emerging regulators of ferroptosis in human diseases. (2025).
\end{itemize}

\paragraph{Meta-Review (for context).}
\emph{Decision: accept; Overall score: 0.75; Rank: 1/4. Justification: robust evidence, actionable insights, and significant therapeutic potential.}

\begin{lstlisting}[style=pyclean,caption={Differential expression, survival, enrichment, and drug-target mining for KIRC ferroptosis genes.}]
import pandas as pd

# Load the RNA-seq dataset for KIRC
data_path = 'data/TCGA.KIRC.sampleMap_HiSeqV2.gz'
kirc_rna_seq = pd.read_csv(data_path, sep='\t', compression='gzip', index_col='sample')

# Load the survival dataset
survival_path = 'data/survival%2FKIRC_survival.txt'
kirc_survival = pd.read_csv(survival_path, sep='\t', index_col='sample')

# Define a list of ferroptosis-related genes based on prior knowledge
ferroptosis_genes = ['SLC7A11', 'GPX4', 'ACSL4', 'ALOX5', 'NCOA4', 'FTH1', 'TFRC']

# Filter the RNA-seq dataset for ferroptosis-related genes
ferroptosis_expression = kirc_rna_seq.loc[kirc_rna_seq.index.intersection(ferroptosis_genes)]

# Subset survival data to match RNA-seq samples
common_samples = list(set(ferroptosis_expression.columns) & set(kirc_survival.index))
ferroptosis_expression = ferroptosis_expression[common_samples]
kirc_survival = kirc_survival.loc[common_samples]

# Perform survival analysis for ferroptosis-related genes
from lifelines import CoxPHFitter
cox_results = []

for gene in ferroptosis_expression.index:
    gene_expression = ferroptosis_expression.loc[gene]
    cox_data = kirc_survival.copy()
    cox_data['expression'] = gene_expression.values

    cph = CoxPHFitter()
    try:
        cph.fit(cox_data, duration_col='OS.time', event_col='OS')
        cox_summary = cph.summary
        cox_summary['Gene'] = gene
        cox_results.append(cox_summary)
    except:
        continue

# Combine results into a single DataFrame
cox_results_df = pd.concat(cox_results)

# Display significant results (p < 0.05)
significant_cox_results = cox_results_df[cox_results_df['p'] < 0.05]
print(significant_cox_results[['Gene', 'exp(coef)', 'p']])
# Observation:
# Status: success
# stdout: Gene  exp(coef)             p
# covariate                                   
# DSS.time      NCOA4   0.989722  1.157488e-42
# expression    NCOA4   0.707844  2.129359e-03
# DSS.time       TFRC   0.989661  2.766492e-43
# DSS.time      ALOX5   0.989670  4.009014e-43
# DSS.time      ACSL4   0.989664  3.164226e-43
# DSS.time    SLC7A11   0.989701  4.187018e-43
# expression  SLC7A11   1.137357  1.313861e-02
# DSS.time       FTH1   0.989677  3.499023e-43
# DSS.time       GPX4   0.989655  3.135233e-43

from gprofiler import GProfiler

# Initialize GProfiler for pathway enrichment analysis
gp = GProfiler(return_dataframe=True)

# Perform pathway enrichment analysis for ferroptosis-related genes
ferroptosis_pathway_enrichment = gp.profile(organism='hsapiens', query=['NCOA4', 'TFRC', 'ALOX5', 'ACSL4', 'SLC7A11', 'FTH1', 'GPX4'])

# Display the top enriched pathways
print(ferroptosis_pathway_enrichment[['source', 'name', 'p_value']].head(10))
# Observation:
# Status: success
# stdout: source                                     name       p_value
# 0   KEGG                              Ferroptosis  2.944772e-12
# 1     WP                              Ferroptosis  9.922609e-11
# 2  GO:BP       negative regulation of ferroptosis  1.896351e-06
# 3  GO:BP                regulation of ferroptosis  4.964507e-06
# 4  GO:BP                              ferroptosis  6.452937e-06
# 5  GO:BP       intracellular iron ion homeostasis  1.172731e-03
# 6  GO:CC                             autolysosome  1.226672e-03
# 7   REAC    Synthesis of 5-eicosatetraenoic acids  2.005590e-03
# 8  GO:CC                       secondary lysosome  3.110110e-03
# 9  GO:BP  long-chain fatty acid metabolic process  4.672125e-03

import pandas as pd

# Load the Probes & Drugs dataset
drug_data_path = 'data/pd_export_01_2025_targets_original.csv'
drug_data = pd.read_csv(drug_data_path, low_memory=False)

# Filter for compounds targeting ferroptosis-related genes
target_genes = ['NCOA4', 'TFRC', 'ALOX5', 'ACSL4', 'SLC7A11', 'FTH1', 'GPX4']
ferroptosis_drugs = drug_data[drug_data['gene_name'].isin(target_genes)]

# Display identified drugs targeting ferroptosis-related genes
print(ferroptosis_drugs[['name', 'gene_name', 'moa']].drop_duplicates())
# Observation:
# Status: success
# stdout: name gene_name        moa
# 1041      PHENOTHIAZINE     ALOX5        NaN
# 1442         Kaempherol     ALOX5        NaN
# 1521         mesalazine     ALOX5  inhibitor
# 1762           ZILEUTON     ALOX5  inhibitor
# 2909    DIALLYL SULFIDE     ALOX5        NaN
# ...                 ...       ...        ...
# 226428         PD215795      GPX4        NaN
# 226538         PD215915      GPX4        NaN
# 226650         PD216127      GPX4        NaN
# 226872         PD216413      GPX4        NaN
# 227058         PD216625      GPX4        NaN
# 
# [380 rows x 3 columns]

cancer_related_moas = ['inhibitor', 'antagonist', 'binder', 'modulator']
ferroptosis_drugs_with_moa = ferroptosis_drugs[ferroptosis_drugs['moa'].notna()]

# Filter for compounds with cancer-related mechanisms of action
prioritized_drugs = ferroptosis_drugs_with_moa[ferroptosis_drugs_with_moa['moa'].str.contains('|'.join(cancer_related_moas), case=False, na=False)]

# Display prioritized drugs
print(prioritized_drugs[['name', 'gene_name', 'moa']].drop_duplicates())
# Observation:
# Status: success
# stdout: name gene_name         moa
# 1521              mesalazine     ALOX5   inhibitor
# 1762                ZILEUTON     ALOX5   inhibitor
# 7212       OLSALAZINE SODIUM     ALOX5   inhibitor
# 9730               DIACEREIN     ALOX5   inhibitor
# 12257             THIMEROSAL   SLC7A11  antagonist
# 12420          SULFASALAZINE   SLC7A11   inhibitor
# 12442          SULFASALAZINE     ALOX5   inhibitor
# 19621             masoprocol     ALOX5   inhibitor
# 19738        Quisqualic acid   SLC7A11   inhibitor
# 21678                BW B70C     ALOX5   inhibitor
# 23301               HONOKIOL     ALOX5   inhibitor
# 34183                  ML162      GPX4   inhibitor
# 46078           MORNIFLUMATE     ALOX5  antagonist
# 46950           TROGLITAZONE     ACSL4   inhibitor
# 50357            MINOCYCLINE     ALOX5   inhibitor
# 50381            Balsalazide     ALOX5   inhibitor
# 50598      MECLOFENAMIC ACID     ALOX5   inhibitor
# 56010          ROSIGLITAZONE     ACSL4   inhibitor
# 56379     DIETHYLCARBAMAZINE     ALOX5   inhibitor
# 57497    AMINOSALICYLIC ACID     ALOX5   inhibitor
# 61579           FOSTAMATINIB     ALOX5   inhibitor
# 64911                  RHEIN     ALOX5   inhibitor
# 89479               PD021959     ALOX5   inhibitor
# 93698               CJ-13610     ALOX5   inhibitor
# 94246             hyperforin     ALOX5   inhibitor
# 95196   MECLOFENAMATE SODIUM     ALOX5   inhibitor
# 98305             PF-4191834     ALOX5   inhibitor
# 110180            ATRELEUTON     ALOX5   inhibitor
# 132469              PD086470     ALOX5   inhibitor
# 139634  BALSALAZIDE DISODIUM     ALOX5   inhibitor
# 180955                  RSL3      GPX4   inhibitor
# 213049              PD194953      FTH1   inhibitor
# 214074             GPX4-IN-5      GPX4   inhibitor

from scipy.stats import ttest_ind

# Define tumor and normal samples
tumor_samples = [col for col in kirc_rna_seq.columns if col.endswith('01')]
normal_samples = [col for col in kirc_rna_seq.columns if col.endswith('11')]

# Perform differential expression analysis for ferroptosis-related genes
diff_expr_results = []
for gene in ['NCOA4', 'TFRC', 'ALOX5', 'ACSL4', 'SLC7A11', 'FTH1', 'GPX4']:
    if gene in kirc_rna_seq.index:
        tumor_expr = kirc_rna_seq.loc[gene, tumor_samples]
        normal_expr = kirc_rna_seq.loc[gene, normal_samples]
        log2_fc = tumor_expr.mean() - normal_expr.mean()
        t_stat, p_val = ttest_ind(tumor_expr, normal_expr, equal_var=False)
        diff_expr_results.append({'Gene': gene, 'Log2_Fold_Change': log2_fc, 'P_Value': p_val})

# Convert results to a DataFrame
diff_expr_df = pd.DataFrame(diff_expr_results)

# Display significant dysregulated genes (p < 0.05)
significant_diff_expr = diff_expr_df[diff_expr_df['P_Value'] < 0.05]
print(significant_diff_expr)
# Observation:
# Status: success
# stdout: Gene  Log2_Fold_Change       P_Value
# 0    NCOA4         -0.414105  8.720619e-21
# 1     TFRC         -0.157482  3.510574e-02
# 2    ALOX5          2.264952  2.440342e-22
# 3    ACSL4         -1.510587  3.482024e-37
# 4  SLC7A11          1.845668  1.789021e-23
# 5     FTH1          0.889489  7.987854e-23
# 6     GPX4          0.471884  1.992841e-09
\end{lstlisting}

\subsection{Case Study 2: ABCC8 and SLC5A2 for PAAD}
\label{case_study_2}
We assessed the novelty of \emph{Targeting ABCC8 and SLC5A2 for Therapeutic Intervention in Pancreatic Adenocarcinoma} via targeted searches on PubMed and Google (keywords: “SLC5A2 pancreatic cancer”). A subsequent study from July 2025 independently confirmed an association between \emph{SLC5A2} (i.e., \emph{SGLT2}) and PAAD \citep{Xie2025}. Contextualizing our findings, prior work had reported prognostic significance for \emph{SGLT1} (but not \emph{SGLT2}) in pancreatic cancer \citep{Du2022}, and most \emph{SGLT2} studies focused on normal pancreatic physiology rather than oncologic roles \citep{Jurczak2011}. Consistent with our protocol in other case studies, we verified that the 2025 confirmation paper was \emph{not} accessed by the agent during generation, supporting that our result is an independent rediscovery that anticipated later literature. In parallel, expression of \emph{ABCC8} has been reported in isolation in the literature \citep{Cervenkova2023}. We also note that a second article (published after the knowledge cut-off) \emph{was} surfaced by the agent at analysis time and reported a correlation for \emph{SLC5A2} in PAAD; the agent correctly cited and used this to refine its conclusions \citep{Yang2024}.

\subsection*{Targeting ABCC8 and SLC5A2 for Therapeutic Intervention in Pancreatic Adenocarcinoma}
\label{sec:abcc8-slc5a2-paad}

\paragraph{Meta-Review (for context).}
\emph{Decision: accept; Overall score: 0.75; Rank: 1/4. Justification: robust computational evidence and actionable insights, making it the most impactful and original submission among its cohort.}

\paragraph{Research Question.}
Can ABCC8 and SLC5A2 serve as actionable therapeutic targets for pancreatic adenocarcinoma (PAAD)?

\paragraph{Hypothesis and Key Findings.}
ABCC8 and SLC5A2 are dysregulated in PAAD and represent promising therapeutic targets. Drugs targeting these genes—\emph{glyburide} (ABCC8 inhibitor) and \emph{canagliflozin} (SLC5A2 inhibitor)—could potentially modulate disease progression.

\paragraph{Rationale / Mechanism.}
ABCC8 has been implicated in multidrug resistance and pancreatic cancer risk, while SLC5A2 is associated with glucose metabolism and pancreatic cancer. Targeting these pathways may disrupt tumor growth and improve therapeutic outcomes.

\paragraph{Empirical Evidence.}
\begin{itemize}
  \item \textbf{Differential expression:} Identified 2{,}531 DEGs in tumor vs.\ normal samples, with \emph{ABCC8} and \emph{SLC5A2} among them.
  \item \textbf{Drug--target landscape:} 5{,}930 drug--target interactions with defined mechanisms of action (MOAs), including \emph{glyburide} (ABCC8 inhibitor) and \emph{canagliflozin} (SLC5A2 inhibitor).
\end{itemize}

\begin{table}[t]
\centering
\small
\setlength{\tabcolsep}{8pt}
\caption{\textbf{Candidate inhibitors targeting ABCC8 and SLC5A2.}}
\label{tab:paad-abcc8-slc5a2-drugs}
\begin{tabular}{lll}
\toprule
\textbf{Target} & \textbf{Inhibitor} & \textbf{Note / MOA} \\
\midrule
ABCC8  & Glyburide     & Sulfonylurea; ABCC8 (SUR1) inhibition \\
SLC5A2 & Canagliflozin & SGLT2 inhibition; glucose transport modulation \\
\bottomrule
\end{tabular}
\end{table}

\paragraph{Literature Evidence.}
\begin{itemize}
  \item ABCC8 is linked to pancreatic cancer risk (OR: 15.058, $P=0.0001$). \emph{Association of glucose-lowering drug target and risk of gastrointestinal cancer: a mendelian randomization study} (2024).
  \item ABCC8 is associated with multidrug resistance in cancer treatment. \emph{Structural bioinformatics studies of six human ABC transporters and their AlphaFold2-predicted water-soluble QTY variants} (2024).
  \item SLC5A2 is connected with pancreatic cancer (OR: 8.096, $P<0.0001$). \emph{Association of glucose-lowering drug target and risk of gastrointestinal cancer: a mendelian randomization study} (2024).
\end{itemize}

\paragraph{Assumptions.}
\begin{itemize}
  \item Dysregulation of \emph{ABCC8} and \emph{SLC5A2} contributes to tumor progression in PAAD.
  \item Drugs targeting these genes are effective in modulating their activity in PAAD.
\end{itemize}

\paragraph{Limitations.}
\begin{itemize}
  \item The functional roles of \emph{ABCC8} and \emph{SLC5A2} in PAAD require further experimental validation.
  \item Potential off-target effects of inhibitors such as \emph{glyburide} and \emph{canagliflozin} need to be assessed.
\end{itemize}

\paragraph{References.}
\begin{itemize}
  \item Association of glucose-lowering drug target and risk of gastrointestinal cancer: a mendelian randomization study. (2024). Copyright © 2024.
  \item Structural bioinformatics studies of six human ABC transporters and their AlphaFold2-predicted water-soluble QTY variants. (2024). Copyright © 2024.
\end{itemize}

\begin{lstlisting}[style=pyclean,caption={Differential expression, survival, enrichment, and drug-target mining for PAAD.}]
import pandas as pd

# Load the gene expression dataset
data_path = "data/TCGA.PAAD.sampleMap%2FHiSeqV2.gz"
gene_expression = pd.read_csv(data_path, sep='\t', compression='gzip', index_col='sample')

# Separate tumor and normal samples
tumor_samples = [col for col in gene_expression.columns if col.endswith("-01")]
normal_samples = [col for col in gene_expression.columns if col.endswith("-11")]

# Calculate mean expression for tumor and normal samples
tumor_mean = gene_expression[tumor_samples].mean(axis=1)
normal_mean = gene_expression[normal_samples].mean(axis=1)

# Calculate log2 fold change (log2FC) and filter for significant DEGs
log2fc = tumor_mean - normal_mean
deg_threshold = 1  # Log2FC threshold for differential expression
degs = log2fc[abs(log2fc) >= deg_threshold]

# Output the number of DEGs and preview the top genes
print(f"Number of DEGs: {len(degs)}")
print(degs.sort_values(ascending=False).head(10))

# Observation:
# Status: success
# stdout: Number of DEGs: 2531
# sample
# CEACAM5     6.291835
# TFF1        6.172998
# UGT1A10     6.144656
# PSCA        5.961392
# AGR2        5.938662
# TFF2        5.777967
# S100P       5.709678
# ERN2        5.639717
# SERPINB5    5.620392
# PITX1       5.555346
# dtype: float64


# Inspect the structure of the enrichment_results dataframe
print(enrichment_results.columns)
print(enrichment_results.head())

# Observation:
# Status: success
# stdout: Index(['source', 'native', 'name', 'p_value', 'significant', 'description',
#        'term_size', 'query_size', 'intersection_size', 'effective_domain_size',
#        'precision', 'recall', 'query', 'parents'],
#       dtype='object')
#   source      native                              name        p_value  \
# 0  GO:CC  GO:0071944                    cell periphery  3.030988e-137   
# 1  GO:CC  GO:0005886                   plasma membrane  2.175693e-127   
# 2  GO:BP  GO:0032501  multicellular organismal process   1.397718e-94   
# 3  GO:BP  GO:0002376             immune system process   2.129839e-81   
# 4  GO:BP  GO:0050896              response to stimulus   2.495028e-73   
# 
#    significant                                        description  term_size  \
# 0         True  "The broad region around and including the pla...       6347   
# 1         True  "The membrane surrounding a cell that separate...       5866   
# 2         True  "Any biological process, occurring at the leve...       7322   
# 3         True  "Any process involved in the development or fu...       2871   
# 4         True  "Any process that results in a change in state...       8999   
# 
#    query_size  intersection_size  effective_domain_size  precision    recall  \
# 0        2123               1132                  22149   0.533208  0.178352   
# 1        2123               1058                  22149   0.498351  0.180361   
# 2        2033               1148                  21026   0.564683  0.156788   
# 3        2033                595                  21026   0.292671  0.207245   
# 4        2033               1265                  21026   0.622233  0.140571   
# 
#      query                   parents  
# 0  query_1              [GO:0110165]  
# 1  query_1  [GO:0016020, GO:0071944]  
# 2  query_1              [GO:0008150]  
# 3  query_1              [GO:0008150]  
# 4  query_1              [GO:0008150]


# Filter results for significant pathways (p-value < 0.05)
significant_pathways = enrichment_results[enrichment_results['p_value'] < 0.05]

# Output the number of significant pathways and the top results
print(f"Number of significant pathways: {len(significant_pathways)}")
print(significant_pathways[['name', 'p_value']].head(10))

# Observation:
# Status: success
# stdout: Number of significant pathways: 1689
#                                   name        p_value
# 0                       cell periphery  3.030988e-137
# 1                      plasma membrane  2.175693e-127
# 2     multicellular organismal process   1.397718e-94
# 3                immune system process   2.129839e-81
# 4                 response to stimulus   2.495028e-73
# 5                             membrane   4.603862e-73
# 6                      immune response   3.509625e-71
# 7  regulation of immune system process   2.033572e-70
# 8                     defense response   2.205079e-68
# 9                      cell activation   2.278886e-67


# Load the drug-target dataset
drug_target_path = "data/pd_export_01_2025_targets_original.csv"
drug_target_data = pd.read_csv(drug_target_path, low_memory=False)

# Extract gene names from the significant pathways
significant_genes = set(deg_genes)

# Filter drug-target interactions for genes in significant pathways
drug_targets = drug_target_data[drug_target_data['gene_name'].isin(significant_genes)]

# Output the number of drug-target interactions and preview the top results
print(f"Number of drug-target interactions: {len(drug_targets)}")
print(drug_targets[['name', 'gene_name', 'moa']].head(10))

# Observation:
# Status: success
# stdout: Number of drug-target interactions: 35929
#                    name gene_name  moa
# 26              UNC1215   L3MBTL4  NaN
# 28              UNC1215    ADRA2C  NaN
# 29              UNC1215     CHRM4  NaN
# 72                PFI-3    ADRA1A  NaN
# 77                PFI-3     NR1I2  NaN
# 89              UNC0642    ADRA1D  NaN
# 141             (+)-JQ1      DNER  NaN
# 142             (+)-JQ1      CCL2  NaN
# 194  DIMETHYL GAMBOGATE    CYP3A4  NaN
# 199  DIMETHYL GAMBOGATE      RORC  NaN


# Filter drug-target interactions for those with a defined MOA
drug_targets_with_moa = drug_targets.dropna(subset=['moa'])

# Prioritize drugs targeting immune-related or membrane-associated genes
immune_related_keywords = ['immune', 'membrane', 'response']
immune_related_drugs = drug_targets_with_moa[
    drug_targets_with_moa['gene_name'].str.contains('|'.join(immune_related_keywords), case=False, na=False)
]

# Output the number of prioritized drug-target interactions and preview the top results
print(f"Number of prioritized drug-target interactions: {len(immune_related_drugs)}")
print(immune_related_drugs[['name', 'gene_name', 'moa']].head(10))

# Observation:
# Status: success
# stdout: Number of prioritized drug-target interactions: 0
# Empty DataFrame
# Columns: [name, gene_name, moa]
# Index: []


# Extract specific pathways related to immune system and membrane
immune_and_membrane_pathways = significant_pathways[
    significant_pathways['name'].str.contains('immune|membrane', case=False)
]

# Extract genes involved in these pathways
immune_and_membrane_genes = set()
for genes in immune_and_membrane_pathways['query']:
    immune_and_membrane_genes.update(genes)

# Filter drug-target interactions for these genes
immune_and_membrane_drugs = drug_targets[drug_targets['gene_name'].isin(immune_and_membrane_genes)]

# Output the number of interactions and preview the top results
print(f"Number of immune and membrane-related drug-target interactions: {len(immune_and_membrane_drugs)}")
print(immune_and_membrane_drugs[['name', 'gene_name', 'moa']].head(10))

# Observation:
# Status: success
# stdout: Number of immune and membrane-related drug-target interactions: 0
# Empty DataFrame
# Columns: [name, gene_name, moa]
# Index: []


# Extract the top significant pathways based on p-value
top_pathways = significant_pathways.nsmallest(10, 'p_value')

# Extract genes involved in these pathways
top_pathway_genes = set()
for genes in top_pathways['query']:
    top_pathway_genes.update(genes)

# Filter drug-target interactions for these genes
top_pathway_drugs = drug_targets[drug_targets['gene_name'].isin(top_pathway_genes)]

# Output the number of interactions and preview the top results
print(f"Number of top pathway-related drug-target interactions: {len(top_pathway_drugs)}")
print(top_pathway_drugs[['name', 'gene_name', 'moa']].head(10))

# Observation:
# Status: success
# stdout: Number of top pathway-related drug-target interactions: 0
# Empty DataFrame
# Columns: [name, gene_name, moa]
# Index: []


# Filter drug-target interactions for any overlap with DEGs
deg_drug_targets = drug_targets[drug_targets['gene_name'].isin(deg_genes)]

# Output the number of DEG-related drug-target interactions and preview the top results
print(f"Number of DEG-related drug-target interactions: {len(deg_drug_targets)}")
print(deg_drug_targets[['name', 'gene_name', 'moa']].head(10))

# Observation:
# Status: success
# stdout: Number of DEG-related drug-target interactions: 35929
#                    name gene_name  moa
# 26              UNC1215   L3MBTL4  NaN
# 28              UNC1215    ADRA2C  NaN
# 29              UNC1215     CHRM4  NaN
# 72                PFI-3    ADRA1A  NaN
# 77                PFI-3     NR1I2  NaN
# 89              UNC0642    ADRA1D  NaN
# 141             (+)-JQ1      DNER  NaN
# 142             (+)-JQ1      CCL2  NaN
# 194  DIMETHYL GAMBOGATE    CYP3A4  NaN
# 199  DIMETHYL GAMBOGATE      RORC  NaN


# Filter for drug-target interactions with a defined MOA
deg_drug_targets_with_moa = deg_drug_targets.dropna(subset=['moa'])

# Output the number of interactions with a defined MOA and preview the top results
print(f"Number of DEG-related drug-target interactions with MOA: {len(deg_drug_targets_with_moa)}")
print(deg_drug_targets_with_moa[['name', 'gene_name', 'moa']].head(10))

# Observation:
# Status: success
# stdout: Number of DEG-related drug-target interactions with MOA: 5930
#             name gene_name         moa
# 415    GLYBURIDE     ABCC8   inhibitor
# 416    GLYBURIDE   SLCO2B1   inhibitor
# 431  DOMPERIDONE      DRD2  antagonist
# 508     LEVODOPA      DRD2     agonist
# 620    AMOXAPINE    ADRA1A  antagonist
# 629    AMOXAPINE      DRD2  antagonist
# 642    AMOXAPINE      HRH1  antagonist
# 648    AMOXAPINE     HTR3A  antagonist
# 673    AMOXAPINE     HTR1B  antagonist
# 700        DPCPX   ADORA2B  antagonist


# Identify genes from the top significant pathways
top_pathway_genes = set()
for genes in top_pathways['query']:
    top_pathway_genes.update(genes)

# Filter for drug-target interactions involving these genes
pathway_drug_targets_with_moa = deg_drug_targets_with_moa[
    deg_drug_targets_with_moa['gene_name'].isin(top_pathway_genes)
]

# Output the number of interactions and preview the top results
print(f"Number of pathway-related drug-target interactions with MOA: {len(pathway_drug_targets_with_moa)}")
print(pathway_drug_targets_with_moa[['name', 'gene_name', 'moa']].head(10))

# Observation:
# Status: success
# stdout: Number of pathway-related drug-target interactions with MOA: 0
# Empty DataFrame
# Columns: [name, gene_name, moa]
# Index: []


# Group drug-target interactions by MOA and count occurrences
moa_counts = deg_drug_targets_with_moa['moa'].value_counts()

# Identify the most common MOAs and their associated drugs and targets
top_moas = moa_counts.head(10).index
top_moa_drug_targets = deg_drug_targets_with_moa[deg_drug_targets_with_moa['moa'].isin(top_moas)]

# Output the most common MOAs and their associated drug-target interactions
print("Most common MOAs and associated drug-target interactions:")
print(top_moa_drug_targets[['name', 'gene_name', 'moa']].head(20))

# Observation:
# Status: success
# stdout: Most common MOAs and associated drug-target interactions:
#                            name gene_name              moa
# 415                   GLYBURIDE     ABCC8        inhibitor
# 416                   GLYBURIDE   SLCO2B1        inhibitor
# 431                 DOMPERIDONE      DRD2       antagonist
# 508                    LEVODOPA      DRD2          agonist
# 620                   AMOXAPINE    ADRA1A       antagonist
# 629                   AMOXAPINE      DRD2       antagonist
# 642                   AMOXAPINE      HRH1       antagonist
# 648                   AMOXAPINE     HTR3A       antagonist
# 673                   AMOXAPINE     HTR1B       antagonist
# 700                       DPCPX   ADORA2B       antagonist
# 765                    EBASTINE      HRH1  inverse agonist
# 811                   CARAZOLOL     ADRB2       antagonist
# 841    CHLORPHENIRAMINE MALEATE      HRH1       antagonist
# 874                 MIRTAZAPINE      HRH1       antagonist
# 875                 MIRTAZAPINE    ADRA2C       antagonist
# 894               DAPAGLIFLOZIN    SLC5A1        inhibitor
# 900   VORTIOXETINE HYDROBROMIDE     HTR3A       antagonist
# 905               CANAGLIFLOZIN    SLC5A1        inhibitor
# 1051        ETHANOLAMINE OLEATE       F12        activator
# 1075                 FOMEPIZOLE     ADH1B        inhibitor
\end{lstlisting}

\subsection{Case Study 3: BIRC5 and PRKD1 in KIRC}
\label{case_study_3}
Science advances not only by discovering new findings but also by \emph{validating} and \emph{reproducing} prior results. In this case study, our agentic system independently recapitulates a published conclusion about \textit{BIRC5} (Survivin) in clear-cell renal cell carcinoma (ccRCC) and extends it with additional analyses and hypotheses around \textit{PRKD1}. Using the TCGA KIRC cohort, our pipeline reaches the same core conclusion as \citet{Wang2021birc} regarding the early diagnostic and prognostic value of \textit{BIRC5}. Because the authors' code was not publicly available, the agent system re-ran the analysis from scratch on TCGA expression and survival endpoints, confirming: (i) \textit{BIRC5} overexpression in tumors relative to normals; and (ii) significant association with adverse outcomes. This strengthens confidence that the signal is robust to implementation details.

The system then expanded the analysis in two directions. Differential pathway enrichment on \textit{BIRC5}-stratified samples highlights reinforcement of cell-cycle programs (e.g., chromosome segregation, mitotic spindle assembly) and mitotic checkpoint activity, consonant with Survivin’s role in chromosomal passenger complexes. Our drug-target mining proposed candidate compounds for follow-up, including Survivin-directed strategies and kinase modulation consistent with the inferred networks. These are hypotheses for experimental testing rather than clinical recommendations.
\textit{PRKD1} is well-studied in renal physiology and polycystic kidney disease \citep{SeegerNukpezah2015}, and has more recently been implicated across cancer-hallmark processes. In KIRC specifically, our co-expression and enrichment analyses suggest that reduced \textit{PRKD1} activity may coincide with dysregulation of nuclear–cytoplasmic transport and broader signaling modules. The joint consideration of \textit{BIRC5} (as an oncogenic driver of mitotic progression) and \textit{PRKD1} (as a putative tumor-suppressive regulator of signaling/export) appears \emph{novel} in the KIRC context and offers a mechanistic basis for complementary intervention hypotheses.

\subsubsection{Therapeutic Targeting of PRKD1 and BIRC5 in Kidney Renal Clear Cell Carcinoma (KIRC): Distinct Pathways and Mechanisms}
\label{sec:prkd1-birc5-kirc}

\paragraph{Meta-Review (for context).}
\emph{Decision: accept; Overall score: 0.75; Rank: 1/4. Justification: robust empirical evidence and actionable insights into distinct pathways.}

\paragraph{Research Question.}
Can PRKD1 and BIRC5 serve as therapeutic targets in KIRC, and what are their distinct biological roles and associated pathways?

\paragraph{Hypothesis and Key Findings.}
PRKD1 and BIRC5 represent promising therapeutic targets in KIRC based on differential expression, survival correlations, and pathway involvement:
\begin{itemize}
  \item \textbf{PRKD1:} Functions as a tumor suppressor; correlates with genes involved in nuclear protein export and cellular signaling.
  \item \textbf{BIRC5:} Acts as an oncogenic driver; correlates with genes enriched in cell cycle processes, including chromosome segregation and mitotic spindle assembly.
\end{itemize}

\paragraph{Rationale / Mechanism.}
\begin{itemize}
  \item \textbf{PRKD1:} Downregulated in tumors; positive survival correlation (Spearman with OS.time $=0.128$, $p=0.0016$). Co-expression network implicates regulation of protein export and cellular signaling.
  \item \textbf{BIRC5:} Upregulated in tumors; negative survival correlation (Spearman with OS.time $=-0.148$, $p=0.0003$). Co-expression network highlights roles in cell-cycle progression and mitosis.
\end{itemize}

\paragraph{Empirical Evidence.}
\begin{itemize}
  \item \textbf{Differential expression:}
  \begin{itemize}
    \item PRKD1: downregulated in tumors (fold change $=-1.178$).
    \item BIRC5: upregulated in tumors (fold change $=2.892$).
  \end{itemize}
  \item \textbf{Survival correlations:}
  \begin{itemize}
    \item PRKD1: positive correlation with OS.time (Spearman $=0.128$, $p=0.0016$).
    \item BIRC5: negative correlation with OS.time (Spearman $=-0.148$, $p=0.0003$).
  \end{itemize}
  \item \textbf{Pathway enrichment:}
  \begin{itemize}
    \item PRKD1: enriched in ``positive regulation of protein export from nucleus'' ($p=0.034$).
    \item BIRC5: enriched in ``chromosome segregation'' ($p=1.25\times10^{-14}$) and ``mitotic spindle assembly'' ($p=4.40\times10^{-10}$).
  \end{itemize}
\end{itemize}

\paragraph{Assumptions.}
\begin{itemize}
  \item PRKD1 and BIRC5 directly influence KIRC progression through their associated pathways.
  \item Identified correlations and pathways reflect biologically relevant mechanisms in KIRC.
\end{itemize}

\paragraph{Limitations.}
\begin{itemize}
  \item Protein-level expression data for PRKD1 and BIRC5 were not available in the RPPA dataset.
  \item KIRC-specific experimental validation of therapeutic interventions targeting these genes is required.
\end{itemize}

\paragraph{Literature and Prior Evidence.}
\begin{itemize}
  \item Abdelrazig et al. (2025). \emph{Protein Kinase D1 (PRKD1) as a Diagnostic, Prognostic, and Immunomodulatory Biomarker in Human Cancers.}
  \item \emph{Survivin as a Multifaceted Oncogenic Driver and Therapeutic Target in Renal Cell Carcinoma.} (2025).
  \item Probes \& Drugs database entries for PRKD1 and BIRC5 inhibitors.
\end{itemize}

\begin{lstlisting}[style=pyclean,caption={Differential expression, survival, enrichment, and drug-target mining for KIRC BIRC5 genes.}]
import pandas as pd
from gprofiler import GProfiler
from scipy.stats import spearmanr

# Load RNA-seq dataset
data_path_rna = 'data/TCGA.KIRC.sampleMap_HiSeqV2.gz'
rna_seq_data = pd.read_csv(data_path_rna, sep='\t', compression='gzip', index_col='sample')

# Load survival dataset
data_path_survival = 'data/survival%2FKIRC_survival.txt'
survival_data = pd.read_csv(data_path_survival, sep='\t', index_col='sample')

# Genes of interest
genes_of_interest = ['PRKD1', 'BIRC5']

# Filter for tumor and normal samples
tumor_samples = [col for col in rna_seq_data.columns if col.endswith('01')]
normal_samples = [col for col in rna_seq_data.columns if col.endswith('11')]

# Calculate mean expression for tumor and normal samples
rna_seq_data['tumor_mean'] = rna_seq_data[tumor_samples].mean(axis=1)
rna_seq_data['normal_mean'] = rna_seq_data[normal_samples].mean(axis=1)

# Compute fold change (log2 ratio)
rna_seq_data['fold_change'] = rna_seq_data['tumor_mean'] - rna_seq_data['normal_mean']

# Extract fold change for genes of interest
fold_changes = rna_seq_data.loc[genes_of_interest, ['tumor_mean', 'normal_mean', 'fold_change']]

# Identify common samples between RNA-seq and survival datasets
common_samples = set(rna_seq_data.columns).intersection(survival_data.index)
common_samples = list(common_samples)

# Merge data for survival analysis
merged_data = survival_data.loc[common_samples].copy()

# Compute survival correlations for genes of interest
correlation_results = {}
for gene in genes_of_interest:
    if gene in rna_seq_data.index:
        merged_data[f'{gene}_expression'] = rna_seq_data.loc[gene, common_samples]
        corr, p_value = spearmanr(merged_data[f'{gene}_expression'], merged_data['OS.time'])
        correlation_results[gene] = (corr, p_value)

# Perform pathway enrichment analysis for genes of interest
gp = GProfiler(return_dataframe=True)
enrichment_results = gp.profile(organism='hsapiens', query=genes_of_interest)

# Display results
fold_changes, correlation_results, enrichment_results.head(10)

# Observation:
# Status: success
# result: (        tumor_mean  normal_mean  fold_change
#  sample                                      
#  PRKD1     8.589648     9.767624    -1.177975
#  BIRC5     5.859713     2.967788     2.891925,
#  {'PRKD1': (np.float64(0.12829828043224845), np.float64(0.001551387689555918)),
#   'BIRC5': (np.float64(-0.14756835893794445),
#    np.float64(0.00026721000654785355))},
#    source      native                                               name  \
#  0     WP   WP:WP1772                 Apoptosis modulation and signaling   
#  1     WP   WP:WP4659                                  Gastrin signaling   
#  2  GO:CC  GO:1990713                                   survivin complex   
#  3  CORUM  CORUM:2580                         Survivin homodimer complex   
#  4  GO:BP  GO:0014723  regulation of skeletal muscle contraction by m...   
#  5  CORUM  CORUM:1117                      CRM1-Survivin mitotic complex   
#  6  CORUM  CORUM:2581                      RasGAP-AURKA-survivin complex   
#  7  CORUM  CORUM:6756                      RasGAP-AURKB-survivin complex   
#  
#      p_value  significant                                        description  \
#  0  0.004547         True                 Apoptosis modulation and signaling   
#  1  0.007313         True                                  Gastrin signaling   
#  2  0.020616         True  "A protein complex that negatively regulates a...   
#  3  0.024966         True                         Survivin homodimer complex   
#  4  0.049542         True  "Any process that modulates the frequency, rat...   
#  5  0.049932         True                      CRM1-Survivin mitotic complex   
#  6  0.049932         True                      RasGAP-AURKA-survivin complex   
#  7  0.049932         True                      RasGAP-AURKB-survivin complex   
#  
#     term_size  query_size  intersection_size  effective_domain_size  precision  \
#  0         90           2                  2                   8752        1.0   
#  1        114           2                  2                   8752        1.0   
#  2          1           2                  1                  22149        0.5   
#  3          1           1                  1                   3383        1.0   
#  4          1           2                  1                  21026        0.5   
#  5          2           1                  1                   3383        1.0   
#  6          2           1                  1                   3383        1.0   
#  7          2           1                  1                   3383        1.0   
#  
#       recall    query          parents  
#  0  0.022222  query_1      [WP:000000]  
#  1  0.017544  query_1      [WP:000000]  
#  2  1.000000  query_1     [GO:0140513]  
#  3  1.000000  query_1  [CORUM:0000000]  
#  4  1.000000  query_1     [GO:0014722]  
#  5  0.500000  query_1  [CORUM:0000000]  
#  6  0.500000  query_1  [CORUM:0000000]  
#  7  0.500000  query_1  [CORUM:0000000]  )


# Load Probes & Drugs dataset
data_path_pd = 'data/pd_export_01_2025_targets_original.csv'
pd_data = pd.read_csv(data_path_pd, low_memory=False, index_col='name')

# Filter Probes & Drugs dataset for PRKD1 and BIRC5
target_genes = ['PRKD1', 'BIRC5']
pd_filtered = pd_data[pd_data['gene_name'].isin(target_genes)]

# Display the filtered compounds and their mechanisms of action
pd_filtered[['gene_name', 'target_name', 'moa']].drop_duplicates()

# Observation:
# Status: success
# result:                               gene_name  \
# name                                      
# NERATINIB                         PRKD1   
# TCS PIM-1 1                       BIRC5   
# GW855857                          PRKD1   
# Bryostatin 1                      PRKD1   
# compound III [PMID: 24080463]     BIRC5   
# 
#                                                                target_name  \
# name                                                                         
# NERATINIB                               Serine/threonine-protein kinase D1   
# TCS PIM-1 1                    Baculoviral IAP repeat-containing protein 5   
# GW855857                                Serine/threonine-protein kinase D1   
# Bryostatin 1                            Serine/threonine-protein kinase D1   
# compound III [PMID: 24080463]  Baculoviral IAP repeat-containing protein 5   
# 
#                                      moa  
# name                                      
# NERATINIB                            NaN  
# TCS PIM-1 1                          NaN  
# GW855857                       inhibitor  
# Bryostatin 1                   activator  
# compound III [PMID: 24080463]  inhibitor  


# Display the first few rows of the RPPA dataset to inspect its structure
rppa_data.head()

# Observation:
# Status: success
# result:                     TCGA-B8-A54D-01  TCGA-G6-A8L7-01  TCGA-B8-A54F-01  \
# sample                                                                  
# 14-3-3_beta-R-V            0.065007        -0.103411        -0.071788   
# 14-3-3_epsilon-M-C        -0.175905         0.130026        -0.080084   
# 14-3-3_zeta-R-V           -0.195639        -0.174381         0.064587   
# 4E-BP1-R-V                -0.286517         1.231338         0.012585   
# 4E-BP1_pS65-R-V           -0.020339         1.542328        -0.325206   
# 
#                     TCGA-B8-A8YJ-01  TCGA-B8-A54K-01  TCGA-3Z-A93Z-01  \
# sample                                                                  
# 14-3-3_beta-R-V            0.556920         0.130937         0.406331   
# 14-3-3_epsilon-M-C         0.175525         0.198440        -0.053131   
# 14-3-3_zeta-R-V           -1.272674         0.168871        -0.321452   
# 4E-BP1-R-V                -0.828272        -0.240631         0.122247   
# 4E-BP1_pS65-R-V           -0.166733         0.063540         0.155350   
# 
#                     TCGA-G6-A8L6-01  TCGA-MW-A4EC-01  TCGA-DV-A4W0-01  \
# sample                                                                  
# 14-3-3_beta-R-V           -0.037139        -0.022034        -0.056487   
# 14-3-3_epsilon-M-C         0.089388         0.027828        -0.089663   
# 14-3-3_zeta-R-V            0.204648        -0.008644         0.013026   
# 4E-BP1-R-V                 0.377911         0.091436         0.014693   
# 4E-BP1_pS65-R-V           -0.000909         0.257222         0.065496   
# 
#                     TCGA-G6-A5PC-01  ...  TCGA-B0-4703-01  TCGA-BP-4981-01  \
# sample                               ...                                     
# 14-3-3_beta-R-V           -0.010247  ...        -0.035964        -0.013955   
# 14-3-3_epsilon-M-C         0.237651  ...        -0.083376         0.030217   
# 14-3-3_zeta-R-V           -0.026489  ...         0.293633         0.267474   
# 4E-BP1-R-V                -0.229184  ...        -0.139995         0.360712   
# 4E-BP1_pS65-R-V            0.608147  ...         0.183363        -0.052082   
# 
#                     TCGA-B8-4622-01  TCGA-B0-4819-01  TCGA-A3-3316-01  \
# sample                                                                  
# 14-3-3_beta-R-V            0.150967         0.025524        -0.047050   
# 14-3-3_epsilon-M-C         0.016982         0.204421        -0.138947   
# 14-3-3_zeta-R-V            0.278651         0.403945        -0.109604   
# 4E-BP1-R-V                 0.041094        -0.425633         0.062104   
# 4E-BP1_pS65-R-V           -0.269524        -0.365388        -0.059234   
# 
#                     TCGA-BP-4347-01  TCGA-B2-5636-01  TCGA-CW-5584-01  \
# sample                                                                  
# 14-3-3_beta-R-V            0.137857        -0.041162        -0.001714   
# 14-3-3_epsilon-M-C        -0.071562         0.087068        -0.112748   
# 14-3-3_zeta-R-V           -0.440944        -0.031086        -0.078699   
# 4E-BP1-R-V                -0.437115        -0.331337         0.282619   
# 4E-BP1_pS65-R-V            0.612101         0.128544        -0.104384   
# 
#                     tumor_mean  normal_mean  
# sample                                       
# 14-3-3_beta-R-V       0.084855          NaN  
# 14-3-3_epsilon-M-C    0.027040          NaN  
# 14-3-3_zeta-R-V       0.039044          NaN  
# 4E-BP1-R-V            0.069128          NaN  
# 4E-BP1_pS65-R-V       0.006978          NaN  
# 
# [5 rows x 480 columns]


# Filter the RPPA dataset for potential aliases or descriptions related to PRKD1 and BIRC5
potential_aliases = ['Protein kinase D1', 'Survivin', 'Baculoviral IAP repeat-containing protein 5']
matching_entries = rppa_data[rppa_data.index.str.contains('|'.join(potential_aliases), case=False)]

# Display matching entries
matching_entries

# Observation:
# Status: success
# result: Empty DataFrame
# Columns: [TCGA-B8-A54D-01, TCGA-G6-A8L7-01, TCGA-B8-A54F-01, TCGA-B8-A8YJ-01, TCGA-B8-A54K-01, TCGA-3Z-A93Z-01, TCGA-G6-A8L6-01, TCGA-MW-A4EC-01, TCGA-DV-A4W0-01, TCGA-G6-A5PC-01, TCGA-B8-A54E-01, TCGA-B8-A54G-01, TCGA-6D-AA2E-01, TCGA-B2-A4SR-01, TCGA-B8-A54H-01, TCGA-MM-A563-01, TCGA-G6-A8L8-01, TCGA-DV-A4VZ-01, TCGA-B8-A54I-01, TCGA-GK-A6C7-01, TCGA-DV-A4VX-01, TCGA-B8-A54J-01, TCGA-MM-A564-01, TCGA-B8-A7U6-01, TCGA-B4-5844-01, TCGA-B0-4701-01, TCGA-BP-4970-01, TCGA-A3-3373-01, TCGA-B0-5113-01, TCGA-B8-5164-01, TCGA-CJ-4878-01, TCGA-BP-5189-01, TCGA-BP-4988-01, TCGA-BP-4351-01, TCGA-BP-4803-01, TCGA-A3-3352-01, TCGA-BP-4965-01, TCGA-BP-4766-01, TCGA-BP-4987-01, TCGA-BP-4787-01, TCGA-B0-5707-01, TCGA-B0-5100-01, TCGA-DV-5573-01, TCGA-BP-4769-01, TCGA-B0-5099-01, TCGA-BP-4959-01, TCGA-CZ-5984-01, TCGA-B0-4852-01, TCGA-CZ-4857-01, TCGA-CZ-4856-01, TCGA-CW-5583-01, TCGA-B8-5163-01, TCGA-BP-5201-01, TCGA-CJ-5686-01, TCGA-BP-4174-01, TCGA-B8-5546-01, TCGA-A3-3306-01, TCGA-B0-4707-01, TCGA-CJ-4642-01, TCGA-A3-3363-01, TCGA-CW-6093-01, TCGA-A3-3324-01, TCGA-BP-5001-01, TCGA-BP-4995-01, TCGA-B0-4693-01, TCGA-A3-3329-01, TCGA-BP-4976-01, TCGA-CJ-4890-01, TCGA-B0-4700-01, TCGA-CJ-5683-01, TCGA-CW-5580-01, TCGA-BP-4756-01, TCGA-B0-5699-01, TCGA-BP-5191-01, TCGA-B0-5098-01, TCGA-A3-3336-01, TCGA-BP-5009-01, TCGA-CW-5589-01, TCGA-CJ-4640-01, TCGA-BP-4334-01, TCGA-B0-4824-01, TCGA-CJ-6030-01, TCGA-B0-4718-01, TCGA-BP-4964-01, TCGA-A3-3376-01, TCGA-B0-5096-01, TCGA-CJ-4638-01, TCGA-B4-5377-01, TCGA-BP-4982-01, TCGA-BP-4968-01, TCGA-B0-5812-01, TCGA-B8-4621-01, TCGA-A3-3387-01, TCGA-BP-4799-01, TCGA-B0-5711-01, TCGA-BP-4355-01, TCGA-CZ-5988-01, TCGA-B0-5108-01, TCGA-CZ-5452-01, TCGA-B0-5094-01, ...]
# Index: []
# 
# [0 rows x 480 columns]


# Broaden search criteria for potential entries related to PRKD1 and BIRC5
broader_aliases = ['kinase', 'survivin', 'apoptosis']
broader_matching_entries = rppa_data[rppa_data.index.str.contains('|'.join(broader_aliases), case=False)]

# Display broader matching entries
broader_matching_entries

# Observation:
# Status: success
# result: Empty DataFrame
# Columns: [TCGA-B8-A54D-01, TCGA-G6-A8L7-01, TCGA-B8-A54F-01, TCGA-B8-A8YJ-01, TCGA-B8-A54K-01, TCGA-3Z-A93Z-01, TCGA-G6-A8L6-01, TCGA-MW-A4EC-01, TCGA-DV-A4W0-01, TCGA-G6-A5PC-01, TCGA-B8-A54E-01, TCGA-B8-A54G-01, TCGA-6D-AA2E-01, TCGA-B2-A4SR-01, TCGA-B8-A54H-01, TCGA-MM-A563-01, TCGA-G6-A8L8-01, TCGA-DV-A4VZ-01, TCGA-B8-A54I-01, TCGA-GK-A6C7-01, TCGA-DV-A4VX-01, TCGA-B8-A54J-01, TCGA-MM-A564-01, TCGA-B8-A7U6-01, TCGA-B4-5844-01, TCGA-B0-4701-01, TCGA-BP-4970-01, TCGA-A3-3373-01, TCGA-B0-5113-01, TCGA-B8-5164-01, TCGA-CJ-4878-01, TCGA-BP-5189-01, TCGA-BP-4988-01, TCGA-BP-4351-01, TCGA-BP-4803-01, TCGA-A3-3352-01, TCGA-BP-4965-01, TCGA-BP-4766-01, TCGA-BP-4987-01, TCGA-BP-4787-01, TCGA-B0-5707-01, TCGA-B0-5100-01, TCGA-DV-5573-01, TCGA-BP-4769-01, TCGA-B0-5099-01, TCGA-BP-4959-01, TCGA-CZ-5984-01, TCGA-B0-4852-01, TCGA-CZ-4857-01, TCGA-CZ-4856-01, TCGA-CW-5583-01, TCGA-B8-5163-01, TCGA-BP-5201-01, TCGA-CJ-5686-01, TCGA-BP-4174-01, TCGA-B8-5546-01, TCGA-A3-3306-01, TCGA-B0-4707-01, TCGA-CJ-4642-01, TCGA-A3-3363-01, TCGA-CW-6093-01, TCGA-A3-3324-01, TCGA-BP-5001-01, TCGA-BP-4995-01, TCGA-B0-4693-01, TCGA-A3-3329-01, TCGA-BP-4976-01, TCGA-CJ-4890-01, TCGA-B0-4700-01, TCGA-CJ-5683-01, TCGA-CW-5580-01, TCGA-BP-4756-01, TCGA-B0-5699-01, TCGA-BP-5191-01, TCGA-B0-5098-01, TCGA-A3-3336-01, TCGA-BP-5009-01, TCGA-CW-5589-01, TCGA-CJ-4640-01, TCGA-BP-4334-01, TCGA-B0-4824-01, TCGA-CJ-6030-01, TCGA-B0-4718-01, TCGA-BP-4964-01, TCGA-A3-3376-01, TCGA-B0-5096-01, TCGA-CJ-4638-01, TCGA-B4-5377-01, TCGA-BP-4982-01, TCGA-BP-4968-01, TCGA-B0-5812-01, TCGA-B8-4621-01, TCGA-A3-3387-01, TCGA-BP-4799-01, TCGA-B0-5711-01, TCGA-BP-4355-01, TCGA-CZ-5988-01, TCGA-B0-5108-01, TCGA-CZ-5452-01, TCGA-B0-5094-01, ...]
# Index: []
# 
# [0 rows x 480 columns]


# Inspect columns of RNA-seq dataset to confirm available tumor sample identifiers
available_tumor_samples = [col for col in rna_seq_data.columns if col.endswith('01')]

# Compare with the original list of tumor samples
missing_samples = set(tumor_samples) - set(available_tumor_samples)

# Display missing samples and refine tumor sample list
missing_samples, available_tumor_samples[:10]  # Display a subset of available samples for verification

# Observation:
# Status: success
# result: ({'TCGA-A3-3336-01', 'TCGA-CJ-4913-01', 'TCGA-CW-6096-01'},
#  ['TCGA-BP-4162-01',
#   'TCGA-DV-5566-01',
#   'TCGA-BP-5191-01',
#   'TCGA-BP-5200-01',
#   'TCGA-BP-4347-01',
#   'TCGA-BP-4770-01',
#   'TCGA-BP-4762-01',
#   'TCGA-BP-4158-01',
#   'TCGA-CJ-4887-01',
#   'TCGA-B2-4098-01'])


# Convert indices of top correlated genes to lists for pathway enrichment
top_correlated_prkd1_list = top_correlated_prkd1.index.tolist()
top_correlated_birc5_list = top_correlated_birc5.index.tolist()

# Perform pathway enrichment analysis for top correlated genes
enrichment_results_prkd1 = gp.profile(organism='hsapiens', query=top_correlated_prkd1_list)
enrichment_results_birc5 = gp.profile(organism='hsapiens', query=top_correlated_birc5_list)

# Display results
top_correlated_prkd1, enrichment_results_prkd1.head(10), top_correlated_birc5, enrichment_results_birc5.head(10)

# Observation:
# Status: success
# result: (          corr_prkd1  corr_birc5
#  PRKD1       1.000000   -0.422963
#  NUMB        0.723506   -0.455379
#  FAM161B     0.656223   -0.393488
#  PPM1A       0.646236   -0.466748
#  L2HGDH      0.643696   -0.476554
#  ALDH6A1     0.643234   -0.507840
#  MOAP1       0.641932   -0.438381
#  RALGAPA1    0.638948   -0.479813
#  GPHN        0.632647   -0.439207
#  FAM179B     0.631081   -0.417649,
#    source      native                                               name  \
#  0  GO:BP  GO:0046827  positive regulation of protein export from nuc...   
#  
#      p_value  significant                                        description  \
#  0  0.034246         True  "Any process that activates or increases the f...   
#  
#     term_size  query_size  intersection_size  effective_domain_size  precision  \
#  0         20           9                  2                  21026   0.222222   
#  
#     recall    query                                           parents  
#  0     0.1  query_1  [GO:0006611, GO:0046824, GO:0046825, GO:0090316]  ,
#         corr_prkd1  corr_birc5
#  BIRC5   -0.422963    1.000000
#  CDC20   -0.398473    0.902944
#  AURKB   -0.445077    0.890122
#  CCNB2   -0.397994    0.885516
#  UBE2C   -0.500449    0.883852
#  HJURP   -0.371927    0.869710
#  MYBL2   -0.433110    0.862108
#  TPX2    -0.356099    0.861416
#  CDCA8   -0.356784    0.859240
#  PTTG1   -0.515681    0.859221,
#    source              native  \
#  0  GO:BP          GO:0007059   
#  1  GO:BP          GO:0098813   
#  2  GO:BP          GO:0000280   
#  3  GO:BP          GO:0048285   
#  4  GO:BP          GO:0051225   
#  5  GO:BP          GO:0051276   
#  6  GO:BP          GO:1901970   
#  7   REAC  REAC:R-HSA-1640170   
#  8  GO:BP          GO:0090307   
#  9  GO:BP          GO:0000070   
#  
#                                                  name       p_value  \
#  0                             chromosome segregation  1.250349e-14   
#  1                     nuclear chromosome segregation  4.892650e-13   
#  2                                   nuclear division  1.034081e-11   
#  3                                  organelle fission  2.579222e-11   
#  4                                   spindle assembly  5.564453e-11   
#  5                            chromosome organization  8.987358e-11   
#  6  positive regulation of mitotic sister chromati...  1.736488e-10   
#  7                                         Cell Cycle  2.065140e-10   
#  8                           mitotic spindle assembly  4.400940e-10   
#  9               mitotic sister chromatid segregation  6.711199e-10   
#  
#     significant                                        description  term_size  \
#  0         True  "The process in which genetic material, in the...        427   
#  1         True  "The process in which genetic material, in the...        323   
#  2         True  "The division of a cell nucleus into two nucle...        452   
#  3         True  "The creation of two or more organelles by div...        500   
#  4         True  "The aggregation, arrangement and bonding toge...        136   
#  5         True  "A process that is carried out at the cellular...        574   
#  6         True  "Any process that activates or increases the f...         21   
#  7         True                                         Cell Cycle        679   
#  8         True  "Mitotic bipolar spindle assembly begins with ...         76   
#  9         True  "The cell cycle process in which replicated ho...        193   
#  
#     query_size  intersection_size  effective_domain_size  precision    recall  \
#  0          10                 10                  21026        1.0  0.023419   
#  1          10                  9                  21026        0.9  0.027864   
#  2          10                  9                  21026        0.9  0.019912   
#  3          10                  9                  21026        0.9  0.018000   
#  4          10                  7                  21026        0.7  0.051471   
#  5          10                  9                  21026        0.9  0.015679   
#  6          10                  5                  21026        0.5  0.238095   
#  7          10                 10                  11004        1.0  0.014728   
#  8          10                  6                  21026        0.6  0.078947   
#  9          10                  7                  21026        0.7  0.036269   
#  
#       query                               parents  
#  0  query_1                          [GO:0022402]  
#  1  query_1                          [GO:0007059]  
#  2  query_1                          [GO:0048285]  
#  3  query_1                          [GO:0006996]  
#  4  query_1  [GO:0007051, GO:0007059, GO:0140694]  
#  5  query_1                          [GO:0006996]  
#  6  query_1  [GO:0010965, GO:0051306, GO:1905820]  
#  7  query_1                        [REAC:0000000]  
#  8  query_1  [GO:0000070, GO:0007052, GO:0051225]  
#  9  query_1  [GO:0000819, GO:0140014, GO:1903047]  )
\end{lstlisting}

\subsection{Meta-review process rejects papers with low novelty or weak evaluation}

\paragraph{Paper:} \emph{Therapeutic Potential of Targeting the PI3K/mTOR Pathway in Kidney Renal Clear Cell Carcinoma (KIRC)}\\
\textbf{Decision:} \emph{Reject} \quad
\textbf{Overall Score:} 0.45 \quad
\textbf{Rank:} 4/4 \\
\textbf{Justification (abridged):} Incremental insights into PI3K/mTOR targeting; modest expression shifts; limited added value over an extensively studied pathway and approved agents.

\medskip
\paragraph{Paper:} \emph{Targeting CDKN2A to Disrupt Oncogene-Induced Senescence and Apoptosis in KIRC}\\
\textbf{Decision:} \emph{Reject} \quad
\textbf{Overall Score:} 0.40 \quad
\textbf{Rank:} 4/4 \\
\textbf{Justification (abridged):} Weak survival evidence and limited mechanistic novelty; CDKN2A/9p21 status is a known prognostic marker in ccRCC, but the work does not convincingly translate this into actionable therapy.

\medskip
\paragraph{Paper:} \emph{Therapeutic Potential of AKT2 in KIRC: Pathway and Drug Target Analysis}\\
\textbf{Decision:} \emph{Reject} \quad
\textbf{Overall Score:} 0.40 \quad
\textbf{Rank:} 4/4 \\
\textbf{Justification (abridged):} Limited novelty and weak survival correlation; evidence for AKT2 as a \emph{specific} ccRCC driver is sparse relative to broader PI3K/AKT/mTOR activation.

\paragraph{Context and expert literature rationale.}
The PI3K/AKT/mTOR axis is long recognized in ccRCC and broadly profiled by TCGA \citep{cancer2013comprehensive}. Clinically, mTOR inhibitors (temsirolimus, everolimus) have shown activity yet modest durability, and have been surpassed in survival by modern standards such as PD-1 blockade and VEGF-targeted TKIs in advanced RCC \citep{Battelli2011,Motzer2015, Motzer2013}. Consequently, papers that merely reiterate PI3K/mTOR “targetability” without new biomarkers, response predictors, or superior combinations add limited novelty. For CDKN2A, deletion at 9p21 is a well-documented adverse prognostic feature in ccRCC \citep{ElMokadem2014}, so proposals centered on its prognostic association—without rigorous causal or translational advances—do not clear the novelty bar. Finally, while AKT pathway activation is frequent in RCC, ccRCC-specific evidence elevating \emph{AKT2} (as distinct from AKT1/AKT3 or upstream PI3K alterations \citep{Guo2015}) is comparatively limited and largely preclinical  making an AKT2-only therapeutic thesis insufficiently substantiated. Taken together, the meta-review rejections are consistent with a mature literature where incremental analyses, weak survival signals, or narrow target rationales fall short of publication standards prioritizing novelty and robust evaluation.

\section{LLM usage}
We used large language models (LLMs) to assist with improving the clarity of writing and refining the formatting of tables and figures. LLMs were not used for research ideation, experimental design, analysis, or any substantive contributions that would merit authorship.
\end{document}